\documentclass{article}

\usepackage{arxiv}

\usepackage[utf8]{inputenc}
\usepackage[T1]{fontenc}
\usepackage{url}
\usepackage{amsfonts}
\usepackage{amssymb}
\usepackage{microtype}
\usepackage{xcolor}
\usepackage[numbers,sort&compress]{natbib}
\usepackage{hyperref}

\usepackage{booktabs}
\usepackage{amsmath}
\usepackage{amsthm}
\usepackage{graphicx}
\usepackage{algorithm}
\usepackage{algpseudocode}
\usepackage{enumitem}

\theoremstyle{plain}
\newtheorem{proposition}{Proposition}
\newtheorem{lemma}{Lemma}[section]

\theoremstyle{definition}
\newtheorem{assumption}{Assumption}

\graphicspath{{figures/}}
\title{The Critic Can Hurt: On the Safety of Critics in LLM-Guided AutoML Search}
\renewcommand{\shorttitle}{The Critic Can Hurt}
\renewcommand{\headeright}{}
\renewcommand{\undertitle}{}
\date{}

\author{%
Gaoyuan Du$^{1}$ \And Xiaoyang Liu$^{1}$ \And Rex Zhou$^{1}$ \And Dengdu Jiang$^{1}$ \And
Rong Fu$^{2}$ \And Meng Ding$^{3}$ \And Defu Cao$^{4}$
\AND
\normalfont $^{1}$Amazon \qquad $^{2}$Independent Researcher \\
\normalfont $^{3}$University of Massachusetts Boston \qquad $^{4}$University of Southern California}

\begin{document}

\maketitle
\suppressfloats[t]

\begin{abstract}
LLM critics increasingly steer iterative search in agentic AutoML, prompt optimization, and
self-refinement---evaluated for whether they help, but rarely for whether they can hurt. We isolate
the critic's marginal effect through a four-arm audit and characterize the loop's tolerance
through controlled fault injection. We show a naive 8B critic drives search below the no-critic
random-search floor on 7 of 8 tabular datasets, losing up to $0.245$ AUC in the worst case; the
magnitude varies strongly with dataset, frontier, and budget. \emph{When}: stronger critics
attenuate the loss on the evaluated suite, without establishing a universal capability threshold.
\emph{Why}: controlled fault injection shows that non-directional errors are nearly costless under
best-so-far search, whereas directional errors consume finite search opportunities; an idealized
effective-sample calculation yields a necessary condition, $q(m{-}1)>\beta$, not a deployment
certificate. \emph{How}: a simple observable validation gate and a learned detector can reduce
realized loss, but neither is guaranteed to preserve the full-budget no-critic baseline. We will
release \textsc{CritFault}, to our knowledge the first benchmark for detecting harmful critic
feedback in AutoML. A critic is not a free addition: audit before trusting, and gate when
capability or verification is insufficient.
\end{abstract}


\begin{figure}[t]
\centering
\includegraphics[width=0.9\linewidth]{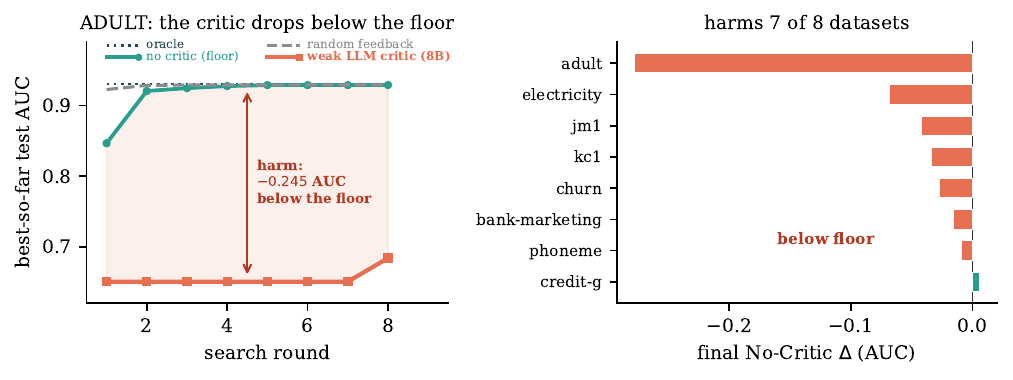}
\caption{\textbf{A weak LLM critic actively harms search.} \textbf{Left:} worst case on
\textsc{adult}---an $8$B critic (coral, AUC $0.68$) pins search far below no-critic random search
(teal, $0.93$) and an independently seeded second random search (gray). \textbf{Right:} the
below-floor sign holds on $7/8$ datasets. Bars show the \emph{No-Critic $\Delta$}, the
critic-guided minus no-critic final test AUC, so negative means below floor (used throughout).}
\label{fig:belowfloor}
\end{figure}

\section{Introduction}
\label{sec:intro}

Automated machine learning (AutoML) searches over preprocessing, feature
transformations, model families, and hyperparameters to build predictive
pipelines with minimal human effort
\cite{thornton2013autoweka,feurer2015autosklearn,olson2016tpot,erickson2020autogluon}.
Recent agentic systems
\cite{chi2024sela,guo2024dsagent,trirat2024automlagent,jiang2025aide}
organize this search as a \emph{propose--evaluate--correct} loop: a proposer
selects a candidate pipeline, the candidate is scored on held-out data, and an
LLM \emph{critic} converts the score and search history into recommendations
that condition the next proposal, while the system retains a best-so-far
incumbent. In principle, critique buys sample efficiency
\cite{hollmann2024caafe}. \textbf{We show that it can do the opposite: a naive
LLM critic can drive search \emph{below} the floor of using no critic at all.}

We study the critic rather than the proposer because the two fail
asymmetrically. A weak proposal is evaluated, scores poorly, and is discarded
by the best-so-far rule; the loop verifies proposals by construction. A
critique receives no such check: it is typically neither scored nor filtered,
yet it conditions every subsequent proposal. Its cost is therefore not a
degraded incumbent---elitist selection protects what has already been
found---but sustained misdirection of a finite evaluation budget away from
configurations that unguided search would otherwise reach.

Evidence from adjacent settings suggests this risk is real but leaves it
unmeasured. LLMs often fail to self-correct without an external signal
\cite{huang2024largelanguage}, reported gains concentrate where a verifier is
available \cite{kamoi2024selfcorrect}, and LLM-as-judge scores are biased and
unstable \cite{zheng2023judgingllm,stureborg2024judgebias}; verbal
self-refinement reports similarly uneven results
\cite{shinn2023reflexion,madaan2023selfrefine}. That evidence, however,
concerns \emph{one-shot} judgment, whose cost is bounded by a single decision.
Closed-loop evaluations, in turn, ablate feedback to measure how much it
\emph{helps} (e.g., SELA's search-policy ablation \cite{chi2024sela});
existing evaluations rarely test the implicit harmlessness assumption directly
against an otherwise identical loop with the critic removed.

To measure that margin, we introduce a \emph{four-arm audit}---\emph{no
critic} (floor), a second independently seeded random search, \emph{real
critic}, and \emph{oracle} (ceiling)---running an identical budget over the
same cached configuration frontier (\S\ref{sec:protocol}). We formalize the underlying
question as \emph{critic safety}: whether, and under what conditions, adding
an LLM critic improves, preserves, or degrades the outcome of an otherwise
identical search. The audit reveals a substantial failure mode. A naive 8B
critic falls below the no-critic floor on 7 of 8 binary tabular datasets
(Fig.~\ref{fig:belowfloor}). Typical losses are ${\sim}0.03$ AUC, reaching
$0.245$ in a deterministic worst case.

Two empirical factors modulate this harm (\S\ref{sec:detect}). First,
\emph{critic capability}: on a targeted six-dataset subset, a 14B critic
improves on the 8B critic on five of six datasets, but stronger critics do not
eliminate negative point estimates (a 70B critic is negative on five of six),
so we find attenuation rather than a capability threshold. Second, the
\emph{search setting}: when the frontier grows from 96 to 250 configurations
and the horizon from 8 to 20 rounds, the same critic's below-floor effect
largely disappears; because both factors change together, we do not attribute
this to headroom alone. The resulting trade-off is asymmetric: on these
frontiers even an ideal critic gains at most ${\sim}0.005$ AUC over the
floor, while the worst-case loss is roughly fifty times larger, and a
practitioner cannot know \emph{ex ante} which regime a given critic and task
fall into.

Why does the loop tolerate some errors and not others? Controlled fault
injection separates error rate from error structure (\S\ref{sec:theory}):
because the loop keeps a best-so-far incumbent, non-directional errors act
like unguided exploration and remain nearly costless even at high rates,
whereas systematically adversarial guidance wastes effective search rounds
and accumulates cost. An idealized effective-horizon analysis condenses this
into the condition $q(m{-}1)>\beta$ (competence $q$, lookahead $m$,
adversarial rate $\beta$), which is necessary for net benefit in the model and
is the leading-order boundary when improvements over the incumbent are rare;
it is not sufficient and does not characterize real critics exactly. Audits in
text, forecasting, retrieval, and a second tabular task are consistent with
the hypothesis that harm tracks mismatch between the critic's priors and the
task.

The condition's one actionable variable is $\beta$. A training-free
validation gate and a learned bad-feedback detector target it without
retraining the critic (\S\ref{sec:prevent}). In deployable end-to-end reruns
that use only observed quantities, the simple gate reduces the raw critic's
average loss relative to uniform search, while neither gate guarantees the
full-budget no-critic baseline on every dataset.

This paper makes three contributions:

\begin{itemize}[leftmargin=1.2em,itemsep=2pt,topsep=2pt,parsep=0pt]

  \item \textbf{Critic safety and a marginal-effect audit (\emph{when}).}
  We formalize \emph{critic safety} and introduce the four-arm audit that
  isolates a critic's marginal effect on a shared frontier
  (\S\ref{sec:protocol}--\S\ref{sec:detect}). The audit uncovers a
  below-floor failure mode that attenuates with critic capability and varies
  with frontier and budget, and observes below-floor results in the published
  CAAFE system (\S\ref{sec:prevent}).

  \item \textbf{An error-structure explanation (\emph{why}).}
  Controlled fault injection and an effective-horizon analysis show the loop
  is resilient to non-directional errors but not to systematically
  adversarial ones, yielding the necessary condition $q(m{-}1)>\beta$ and
  its tail approximation (\S\ref{sec:theory}). Cross-domain audits are
  consistent with its qualitative predictions.

  \item \textbf{Prevention and a benchmark (\emph{how}).}
  A validation gate and a learned bad-feedback detector target $\beta$
  using only observed quantities and are evaluated end to end
  (\S\ref{sec:prevent}). We also contribute \textsc{CritFault}, to our
  knowledge the first benchmark for detecting harmful critic feedback in
  tabular AutoML, to be released with all artifacts upon publication.

\end{itemize}

An LLM critic is thus not a free addition but a component whose marginal
effect must be measured. Before trusting one, audit it against a no-critic
control at the intended budget; when critic capability or verifier reliability is
insufficient, gate or omit the critique.

\section{Related Work}
\label{sec:related}

\textbf{Classical and LLM-based AutoML.} Classical pipeline optimizers
\cite{feurer2015autosklearn,feurer2022autosklearn2,thornton2013autoweka,olson2016tpot,ledell2020h2o,erickson2020autogluon}
search configurations via Bayesian, evolutionary, or ensembling methods, with sample-efficient
hyperparameter tuning \cite{hutter2011smac,falkner2018bohb} and standardized suites
\cite{gijsbers2019amlb,vanschoren2014openml}; strong tabular learners
\cite{chen2016xgboost,ke2017lightgbm,hollmann2023tabpfn,hollmann2025tabpfnv2,grinsztajn2022tabular}
set the ceiling they chase. These optimize a validation metric directly and never reason about
\emph{why} a configuration failed. More recent systems add an LLM to drive pipeline construction:
LLM-orchestrated search with targeted refinement \cite{zhang2023automlgpt,chi2024sela,nam2025mlestar},
retrieval/planning with iterative execution \cite{guo2024dsagent,hong2024datainterpreter}, search
over plans or code \cite{trirat2024automlagent,jiang2025aide}, and RL-trained MLE agents
\cite{liu2025mlagent,yang2025rlmle}; LLMs also serve as feature engineers \cite{hollmann2024caafe}
or direct predictors \cite{manikandan2023weak}, evaluated end-to-end by agentic benchmarks
\cite{chan2024mlebench,huang2024mlagentbench,qiang2025mledojo}. All use the LLM primarily as a
\emph{proposer}, assessed at the outcome level; our object is the complementary \emph{critic}
role---the \emph{conditions} under which an added critic turns from helpful to harmful, which these
works do not isolate.

\noindent\textbf{Verbal feedback and self-critique.} Reflexion \cite{shinn2023reflexion}, Self-Refine
\cite{madaan2023selfrefine}, ReAct \cite{yao2023react}, self-correction training
\cite{welleck2023selfcorrect}, and trained critics \cite{mcaleese2024critics} show verbal critique
can help on coding/reasoning, while a consistent line of negative results shows intrinsic
self-correction can be net-\emph{negative}: it degrades accuracy once oracle labels are removed
\cite{huang2024largelanguage,kamoi2024selfcorrect}, often turns correct answers
incorrect~\cite{tie2025correctbench}, and self-verification is itself unreliable
\cite{stechly2024selfverification}. Closest to our thesis, an imperfect in-context evaluator can
induce \emph{reward hacking} \cite{pan2024rewardhacking}---a proxy-over-optimization inherent to
inference-time selection such as best-of-$N$ \cite{khalaf2025inferencetime}, with
\emph{self-preference} a concrete driver \cite{panickssery2024selfpreference}. This evidence is
mostly on math/code with intrinsic self-evaluation; we study an \emph{externally anchored} critic on
tabular data, sharpening ``fails without external grounding'' into a measured harm \emph{below the
do-nothing floor} with an idealized necessary condition for when a critic can help.

\noindent\textbf{LLM-as-judge reliability and verifier-gated selection.} A growing literature documents biases
and instability in LLM-as-judge setups \cite{zheng2023judgingllm,stureborg2024judgebias,zheng2024largejudge},
while trained \emph{verifiers} \cite{cobbe2021verifiers,lightman2024letsverify,zhang2025genverifiers},
self-taught evaluators \cite{wang2024selftaught}, and prover--verifier games
\cite{kirchner2024proververifier} gate generation. Gating on verifier reliability is now an active
theme \citep{saadfalcon2025weaver,kiyani2026trust,kwok2026verifier}, but these operate as candidate
reranking or task-level verification. Our Verification-Gated Critique differs in setting and
mechanism: it gates \emph{online, inside} the search loop from observed validation utilities,
motivated by our net-benefit condition. \citet{itkin2026delayed}
similarly show that \emph{delayed} verification destabilizes multi-agent belief---independent
evidence that verification, wrongly placed, can do harm. Relatedly, the generation--verification gap
widens with compute \citep{song2024mindgap,davidson2026facts}; the capability attenuation we
observe (\S\ref{sec:detect}) is a consistent but distinct phenomenon, measuring net harm to an external
search rather than intrinsic self-improvement.

\noindent\textbf{Noise tolerance in feedback-driven optimization.} Robust preference optimization
gives explicit noise thresholds---rDPO \citep{chowdhury2024rdpo} tolerates flip rates below $1/2$,
symmetric losses add robustness \citep{nishimori2025sympo}. Our resilience threshold
(\S\ref{sec:theory}) is structurally different: it arises from \emph{elitist best-so-far} dynamics
rather than gradient preference training, and decomposes bad feedback into random (nearly free)
versus systematically adversarial (costly)---a distinction a symmetric flip-rate model cannot
capture.

\section{Problem Setup and Evaluation Protocol}
\label{sec:protocol}

We ask whether feedback from an LLM critic improves---or degrades---finite-budget
search over a fixed space of tabular ML pipelines. This section defines the
search loop and the two instruments the rest of the paper builds on: a
\emph{four-arm audit} that measures a real critic's end-to-end effect
(\S\ref{sec:detect}) and \emph{controlled fault injection} that varies
feedback-error rate and structure independently, motivating the theory of
\S\ref{sec:theory}.

\begin{figure}[t]
\centering
\includegraphics[trim={0 0 241pt 0},clip,width=0.5\linewidth]{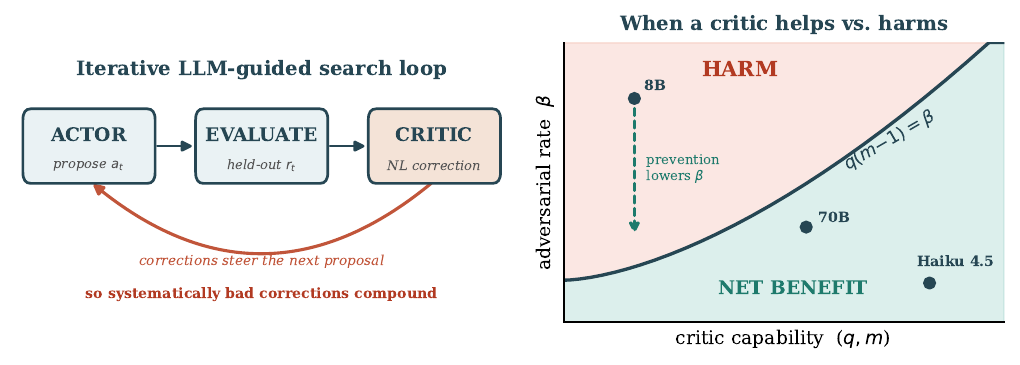}
\caption{The iterative LLM-guided search loop we study.}
\label{fig:overview}
\end{figure}

\subsection{Critic-guided search}
\label{subsec:setup}
\label{subsec:loop}

A critic-guided AutoML agent tunes a tabular pipeline over $T$ rounds
(Fig.~\ref{fig:overview}; Alg.~\ref{alg:loop} in App.~\ref{app:impl}).
Each round, a \emph{proposer} selects an untried configuration $a_t$ from a
finite space $\mathcal{A}$---each $a$ a complete pipeline (model family,
feature-transformation set, hyperparameter preset)---which is trained and
scored on a frozen validation split; an LLM \emph{critic} then turns the score
and history into a short natural-language correction (e.g., \emph{``boosting
underperforms here; try a neural family next''}) conditioning the next
proposal. The agent returns the best-so-far validation incumbent $\hat a_T$,
whose quality is the hidden test utility $s_{\mathrm{test}}(\hat a_T)$. For
dataset $D$ and condition $r$, we write
$J_D(r)=\mathbb{E}[s_{\mathrm{test}}(\hat a_T^{(r)})]$, the expected test
utility over search randomness. Systems such as SELA
\cite{chi2024sela} and AIDE \cite{jiang2025aide} instantiate this
propose--evaluate--correct loop on the premise that critique makes search more
sample-efficient; we ask whether confidently-wrong critique can instead steer
search \emph{below} unguided exploration.

Two design choices isolate the quantity of interest: the critic acts
\emph{only} through the feedback channel (it never trains or executes code,
alters the candidate space, or sees test utilities), so any effect is
attributable to its guidance rather than tool-use failures; and the object of
evaluation is therefore the critique's effect on the full trajectory, not the
correctness of any individual critique. Formal history and feedback maps are
deferred to App.~\ref{app:impl}.

\subsection{Instantiation and the shared cached frontier}
\label{subsec:frontier}

\paragraph{Data flow.}
Each dataset is partitioned once into train/validation/test splits, and every
configuration in a 96-configuration pipeline space $\mathcal{A}$ (design in
App.~\ref{app:impl}) is scored \emph{before} search, with validation and test
AUC cached. During search, selecting a configuration reveals only its cached
validation AUC; test AUC stays hidden and evaluates only the final incumbent.
Because all conditions on a dataset share this same frozen frontier---candidate
space, split, budget, outcomes---they differ only in the trajectory-selecting
mechanism, isolating search-policy differences from training randomness,
execution failures, and pool differences (access rules in
App.~\ref{app:formal-setup}).

\paragraph{Critic and proposer.}
In the experiments reported here, critique and proposal are realized by a
single LLM call per round. The LLM receives the evaluated history
(configuration names and their validation AUCs) and the menu of untried
configurations, states its assessment, and names the next configuration. If
the response names no untried configuration, the next configuration is drawn
uniformly from the menu (a fallback). Without feedback the loop reduces to
uniform sampling. The audited object is therefore this feedback-conditioned
LLM selection policy as a whole, including its parser and fallback; we do not
separate the effect of the critique from that of the mapping to a
configuration. The primary
naive critic is Llama-3.1-8B; the capability analysis of \S\ref{sec:detect}
evaluates seven tiers (3B--70B open-weight models and Claude Haiku 4.5)
across four providers, all with deterministic decoding through a
provider-agnostic interface (identifiers and settings in
App.~\ref{app:impl}).

\subsection{Four-arm audit and controlled fault injection}
\label{subsec:audit}

\paragraph{Four-arm critic-safety audit.}
We compare four budget-matched conditions on the shared frontier that differ
only in their feedback or selection mechanism:
(i)~\emph{no critic}---no critique ($c_t=\varnothing$), uniform sampling over
untried configurations; the unguided-search floor;
(ii)~\emph{random feedback}---in the reported runs, an independently seeded
uniform random search that does not pass feedback through the LLM; it is a
second draw of the floor, not a semantic placebo;
(iii)~\emph{real critic}---feedback from the LLM under evaluation, accepted
without verification unless stated; and
(iv)~\emph{oracle}---bypasses the language channel and directly selects the
highest-validation-utility untried configuration; an unattainable ceiling.
We report two paired contrasts,
\[
\begin{aligned}
\Delta_{\mathrm{NC}}(D)&=J_D(\mathrm{real})-J_D(\mathrm{no\text{-}critic}),\\
\Delta_{\mathrm{RF}}(D)&=J_D(\mathrm{real})-J_D(\mathrm{random\text{-}feedback}),
\end{aligned}
\]
where $\Delta_{\mathrm{NC}}$ measures the system-level value of critic
guidance over unguided search. Because arm (ii) is a random search,
$\Delta_{\mathrm{RF}}$ mirrors $\Delta_{\mathrm{NC}}$ under a different seed and
does not isolate the value of critique content; all conclusions below rest on
$\Delta_{\mathrm{NC}}$. We
say a critic is \emph{below floor} when $\Delta_{\mathrm{NC}}(D)<0$;
prevalence statements (``7 of 8 datasets'') refer to signs of dataset-level
point estimates, while the headline inferential claim rests on a suite-level
sign-flip permutation test, with per-dataset bootstrap intervals reported
alongside (details in App.~\ref{app:impl}). Because all outcomes are cached,
auditing an additional critic requires no repeated pipeline training.

\paragraph{Controlled fault injection.}
The audit cannot separate how \emph{often} a critic errs from how its errors
are \emph{structured}, entangled in a real trajectory. We therefore corrupt a
clean reference policy $\pi^\star$: each round follows $\pi^\star$ with
probability $1-p$ and a corrupted choice with probability $p$, sweeping $p$.
Three structures span the error space---\textsc{Random} (uniform untried;
non-directional), \textsc{Worst} (lowest-utility untried; globally
adversarial), and \textsc{Mislead} (lowest-utility incumbent neighbor; locally
plausible but directionally harmful). These are diagnostic stress tests with
privileged frontier information, not deployable policies; they realize the
limiting error structures of \S\ref{sec:theory}. Degradation curves and the
resilience threshold are defined in App.~\ref{app:impl}.

\subsection{Datasets and experimental scope}
\label{subsec:data}

We evaluate on 12 public OpenML classification datasets spanning sample size,
dimensionality, class imbalance, and domain (Table~\ref{tab:datasets}). We
report in-space headroom
$h_D=J_D(\mathrm{oracle})-J_D(\mathrm{no\text{-}critic})$ (AUC, $T{=}8$,
96-configuration frontier) as a continuous quantity: it is below $0.01$ on all
12 datasets, ranging from $0.0013$ (\textsc{adult}) to $0.0077$
(\textsc{electricity}) on the eight binary datasets and \textsc{letter}, and
$\le 0.0003$ on \textsc{mfeat-factors}, \textsc{segment}, and \textsc{splice}.
We refer to the latter three as the \emph{near-zero-headroom} datasets; this is
a descriptive grouping, not a pre-specified threshold. Each analysis uses an
explicitly defined subset (Table~\ref{tab:analysis-scope}): the audit the
eight binary datasets, fault injection the other nine, the capability ladder a
targeted six,
and \textsc{CritFault} all twelve. Unless stated, the primary audit uses $T=8$,
eight matched seeds, and deterministic decoding; exact settings in
App.~\ref{app:impl}.

\section{\emph{When}: Capability and Search Setting}
\label{sec:detect}

We now characterize when critic guidance becomes harmful and when below-floor
harm disappears. We examine two empirical factors: the search setting
(frontier size and horizon) and the capability of the critic. All experiments use the shared
configuration frontiers and evaluation protocol of \S\ref{sec:protocol}.

\begin{figure}[t]
\centering
\includegraphics[width=0.9\linewidth]{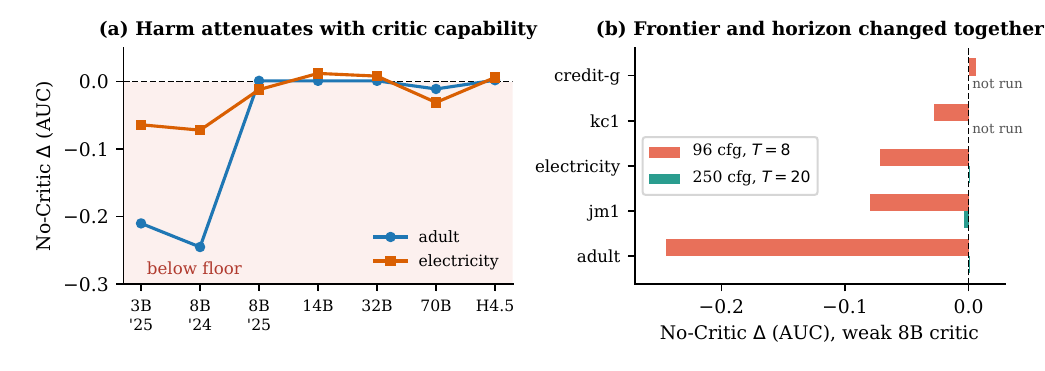}
\caption{\textbf{\emph{When}.} \textbf{(a)}~Capability ladder
(No-Critic $\Delta$, 8 seeds): \textsc{adult} falls from $-0.245$ (2024-gen
8B) to $|\Delta|\le0.011$ at every $\ge$14B tier (Table~\ref{tab:ladder6}).
\textbf{(b)}~The same 8B critic on the 96-configuration frontier with $T{=}8$
(coral) and, for three datasets, on a 250-configuration frontier with
$T{=}20$ (teal); frontier and horizon change together, so the contrast does
not isolate either factor.}
\label{fig:axis1}
\end{figure}

\subsection{Reference policies establish limited upside}
\label{subsec:upside}

We first quantify the improvement available from informative guidance. The
clean hill-climbing reference achieves a positive but small No-Critic delta on
each of the nine datasets outside the near-zero-headroom group ($+0.001$ to
$+0.005$ AUC over 30 seeds, Table~\ref{tab:e3}; suite-level Wilcoxon
$p=0.001$). This limited upside is not a weak search space: the in-space oracle
matches or beats AutoGluon-Tabular on 5 of 8 datasets (Table~\ref{tab:sota}),
so the floors we measure harm against are SOTA-competitive.

In-space headroom is small on every dataset ($0$ to $0.0077$ AUC,
\S\ref{subsec:data}); on the three near-zero-headroom datasets the oracle and
the unguided-search reference are nearly indistinguishable. The reference
results therefore establish an asymmetric opportunity: the attainable upside of guidance is
small on these fixed frontiers, whereas the downside of systematically
misleading guidance can be considerably larger.

\subsection{Weak critics can fall below the search floor}
\label{subsec:belowfloor}

Replacing the clean reference with a real LLM critic changes the outcome
substantially. A naive Llama-3.1-8B critic (temperature 0) has a negative
No-Critic delta on 7 of the 8 binary datasets, ranging from $-0.009$ AUC on
\textsc{phoneme} to $-0.245$ on \textsc{adult}; only \textsc{credit-g} has a
positive point estimate ($+0.006$; Fig.~\ref{fig:belowfloor}). In the
representative \textsc{adult} trajectory the critic emits fluent rationales
while its picks stay below the incumbent (App.~\ref{app:cases};
\S\ref{subsec:mechanism}).

An exact one-sided sign-flip test over all $2^8$ sign assignments
(\S\ref{subsec:audit}) gives $p=0.0078$ for a negative suite-level mean, so
the negative direction is not driven by a single task. It persists
under stochastic decoding and additional seeds, though the largest magnitudes
occur under deterministic decoding; per-run logs are released with the code.

\subsection{Harm attenuates with the search setting and critic capability}
\label{subsec:attenuate}

\paragraph{Harm attenuates on a larger frontier with a longer horizon.}
We evaluate the same 8B critic on an expanded 250-configuration frontier with a
longer horizon of $T{=}20$. In this joint setting the below-floor effect
decreases by one to two orders of magnitude:
\textsc{adult} $-0.245\!\to\!+0.0003$, \textsc{electricity}
$-0.072\!\to\!+0.0003$, \textsc{jm1} $-0.080\!\to\!-0.004$. Because frontier
and horizon change together, this contrast does not identify whether headroom,
budget, or their interaction drives the attenuation. A separate stress test
shows that enlarging the space does not mechanically reduce harm for a
systematically adversarial critic and can amplify it
(Table~\ref{tab:spacesize}).

\paragraph{Harm attenuates with critic capability.}
We repeat the real-critic arm across a seven-tier capability ladder over four providers and two
generations on six datasets (Table~\ref{tab:ladder6}; Fig.~\ref{fig:axis1}a). The weak 2024 8B
critic is negative on 5 of 6 datasets (up to $0.245$ AUC); at the 14B tier every point estimate is
within $0.012$ of zero, although three of six remain negative. Relative to the 8B critic, the 14B
critic improves on five of six datasets (exact one-sided paired sign-flip test, $p=0.047$). The
pattern is not monotone in size or release date: a 2025 3B critic stays harmful on 3 of 4
(\textsc{adult} $-0.210$), and Llama-3.3-70B is negative on five of six datasets (up to $-0.031$).
Because these six datasets were chosen partly for the 8B critic's effect (App.~\ref{app:impl}),
the ladder supports task-dependent attenuation, not a capability threshold.

\begin{table}[t]
\centering
\caption{Capability ladder: No-Critic $\Delta$ (AUC, 8 seeds) across seven
tiers spanning four providers and two generations (2024--2025). The 3B model
was evaluated on four of the six datasets.}
\label{tab:ladder6}
\footnotesize
\setlength{\tabcolsep}{3.5pt}
\begin{tabular}{l@{\hskip 5pt}rrrrrrr}
\toprule
 & 3B$^{25}$ & 8B$^{24}$ & 8B$^{25}$ & 14B & 32B & 70B & H4.5 \\
\midrule
adult       & $-.210$ & $-.245$ & $+.001$ & $+.001$ & $+.001$ & $-.011$ & $+.002$ \\
electr.     & $-.064$ & $-.072$ & $-.012$ & $+.012$ & $+.008$ & $-.031$ & $+.006$ \\
jm1         & $-.008$ & $-.080$ & $-.003$ & $-.002$ & $-.000$ & $-.011$ & $+.002$ \\
kc1         & $+.000$ & $-.028$ & $-.000$ & $-.001$ & $-.001$ & $-.023$ & $-.000$ \\
bank-mkt    &         & $-.007$ & $+.000$ & $+.000$ & $+.001$ & $-.007$ & $+.001$ \\
credit-g    &         & $+.006$ & $-.005$ & $-.012$ & $-.007$ & $+.006$ & $-.001$ \\
\bottomrule
\end{tabular}\\[2pt]
\scriptsize 3B$^{25}${=}Ministral-3B (Mistral '25); 8B$^{24}${=}Llama-3.1
(Meta '24); 8B$^{25}${=}Ministral-3-8B (Mistral '25); 14B{=}Ministral-14B;
32B{=}Qwen3; 70B{=}Llama-3.3; H4.5{=}Haiku\,4.5 (Anthropic).
\end{table}

\paragraph{Robustness.}
The harm is neither a parsing-failure artifact (harm and parse-failure are anti-correlated across
the suite; App.~\ref{app:additional}) nor a naive-prompting one---chain-of-thought and
self-consistency mitigate but do not eliminate it---a point we develop when motivating prevention
(\S\ref{subsec:motivation}).

\subsection{Magnitude and operational consequences}
\label{subsec:magnitude}

The harm is asymmetric relative to the available upside. On the fixed
frontier, the largest gain from clean reference guidance is approximately
$0.005$ AUC (Table~\ref{tab:e3}), whereas the naive 8B critic loses
approximately $0.03$ AUC on a typical harmed dataset and $0.245$--$0.28$ in
the worst case (Fig.~\ref{fig:axis1}a)---roughly six times the largest observed
reference-policy gain typically, and up to fifty times in the worst case.

Below-floor harm is thus largest for a weak critic on these small, fixed
frontiers at short horizons, and because upside is bounded while downside is
not, the risk is asymmetric. The next section asks why the loop tolerates some
critic errors but not others.

\section{\emph{Why}: Error Structure, Not Error Rate}
\label{sec:theory}

The capability results identify when below-floor harm is likely, but not why some critic errors
accumulate while others do not. We isolate error structure through controlled fault injection and
derive its consequences from the loop's best-so-far dynamics. The intuition: the loop reports the
best configuration seen so far, so a random error merely spends a round like the no-critic loop and
cannot lower the incumbent, whereas a systematic error that steers away from the high-value region
denies the incumbent its chance to improve round after round. A critic can thus err often yet stay
harmless, or err rarely but directionally and do real damage. We make this precise across four
levels of claim: exact structural results, a probabilistic bound, a first-order criterion, and
empirical tests.

\subsection{Fault injection reveals a wide but asymmetric safety envelope}
\label{subsec:envelope}

The controlled interventions of \S\ref{subsec:audit} reveal two empirical regularities
(Table~\ref{tab:fault}, 30 seeds). First, \textsc{random} corruption is nearly costless: even at
$p{=}1$, final degradation is below $0.005$ AUC on every tested dataset. Second, systematically
adversarial corruption (\textsc{worst}/\textsc{mislead}) exhibits a broad but finite resilience
region: the measured threshold is $p^\star{=}0.6$--$0.9$ on the primary binary datasets, with
\textsc{letter} at the boundary value $1.0$. Error frequency alone therefore does not determine
harm: errors at similar rates differ sharply depending on whether they preserve exploration or
systematically redirect it---unlike the symmetric flip-rate thresholds of robust preference
optimization (rDPO's $\varepsilon<1/2$ \citep{chowdhury2024rdpo}). We explain both regularities from
the loop's structure.

\paragraph{Model.}
Let $\mathcal{A}=\{a_1,\dots,a_N\}$ be a finite configuration space with fixed cached scores
$s(a)\in[0,1]$; a $T$-round run selects without replacement and reports the incumbent
$b_T=\max_{t\le T}s(a_{\pi(t)})$. The elitist update (Alg.~\ref{alg:loop}) gives the structural
fact we exploit:
\begin{equation}
  b_t \;=\; \max(b_{t-1},\, s(a_{\pi(t)})) \;\ge\; b_{t-1}
  \qquad\text{(monotone, ``elitist'').}
  \label{eq:mono}
\end{equation}
A corruption process replaces the selected configuration with probability $p$, independently across
rounds: \textsc{random} substitutes a uniform untried configuration, \textsc{adversarial}
(\textsc{worst}/\textsc{mislead}) an argmin-type one. The no-critic floor selects uniformly at
random every round.

\subsection{Best-so-far search and random corruption}
\label{subsec:randomfree}

\begin{proposition}[Uniform corruption preserves the random-search floor]
\label{prop:random}
Suppose every selected configuration is uniformly distributed over the remaining untried
configurations, regardless of whether the round is labeled clean or corrupted. Then for every
$p\in[0,1]$ and horizon $T$,
\[
  \mathbb{E}\big[b_T^{\mathrm{uniform}(p)}\big] \;=\; \mathbb{E}\big[b_T^{\text{no-critic}}\big].
\]
\end{proposition}
\noindent The proof (App.~\ref{app:proofs}) follows from without-replacement exchangeability:
both loops induce the same distribution over the visited set, on which $b_T$ depends alone.

This is an exact floor-preservation result for the uninformative regime a naive critic degenerates
to; it does not claim that randomizing an \emph{informative} critic is costless. It explains why
non-directional corruption barely degrades search (random-mode cost within noise of zero on all 12
datasets, Table~\ref{tab:fault}): random selections still explore the same pool and cannot lower an
observed incumbent. Falling below the random-search floor therefore requires selection that
departs from uniform in a harmful direction; the controlled \textsc{worst}/\textsc{mislead}
interventions show that such directional error is sufficient to do so.

\subsection{Adversarial corruption and the resilience threshold}
\label{subsec:adversarial}

For adversarial corruption we need only that a corrupted round contributes no improvement.
\begin{assumption}[Wasted adversarial rounds]
\label{ass:waste}
An adversarial round selects a configuration $a$ with $s(a)\le b_{t-1}$, i.e.\ it never raises the
incumbent. (\textsc{worst} picks the global argmin among untried; \textsc{mislead} the worst
neighbor of the incumbent; both satisfy this once $b_{t-1}$ exceeds the low-score region.)
\end{assumption}
\noindent Empirically $90.5\%$ of adversarial picks fail to raise the incumbent (per-dataset
$87$--$93\%$), so the assumption holds for most rounds and the bound degrades gracefully otherwise.
Let $T_{\mathrm{sat}}$ be the \emph{saturation budget}: the number of effective rounds after which
the clean loop is within $\delta$ of its ceiling ($\mathbb{E}[b_{T_{\mathrm{sat}}}^{\text{clean}}]
\ge b^\star-\delta$, $b^\star=\max_a s(a)$); empirically $T_{\mathrm{sat}}\!\approx\!1$--$3$
(\S\ref{subsec:upside}).

\begin{proposition}[Resilience under adversarial corruption]
\label{prop:threshold}
Let $K\sim\mathrm{Binomial}(T,1-p)$ be the number of effective rounds. Under
Assumption~\ref{ass:waste}, if the clean loop is within $\delta$ of its ceiling whenever
$K\ge T_{\mathrm{sat}}$, then, since scores lie in $[0,1]$,
\[
  \mathbb{E}\,[\,b^\star - b_T\,] \;\le\; \delta \;+\; \Pr\!\big[K < T_{\mathrm{sat}}\big].
\]
Consequently, for a tolerated additional failure probability $\epsilon$, the loop is
$(\delta+\epsilon)$-robust whenever
$\Pr[\mathrm{Binomial}(T,1-p)<T_{\mathrm{sat}}]\le\epsilon$, and we write $p^\star(\epsilon)$ for
the largest such $p$. The simpler expression
\[
  p^\star \;\approx\; 1 - \frac{T_{\mathrm{sat}}}{T}
\]
is the expectation-level approximation to this probabilistic threshold.
\end{proposition}
\noindent The proof (App.~\ref{app:proofs}) splits on $\{K\ge T_{\mathrm{sat}}\}$, using that
corrupted rounds leave the monotone incumbent unchanged. With $T{=}8$ and $T_{\mathrm{sat}}\in[1,3]$
the approximation predicts $p^\star\in[0.63,0.88]$, matching the measured $0.6$--$0.9$
(Table~\ref{tab:fault}): the loop is robust not because the critic is accurate, but because elitist
monotonicity converts its job from never erring into supplying $T_{\mathrm{sat}}$ good rounds out of
$T$.

\subsection{The adversarial signature in real critics}
\label{subsec:mechanism}

Before building a criterion on the adversarial rate, we examine whether \emph{real} critics show
the systematically adversarial structure of Assumption~\ref{ass:waste}. These diagnostics are
consistent with directional mismatch but do not show that every real-critic step is adversarial.
First,
\emph{family-selection divergence}: harmed trajectories under-select the families that dominate the
cached frontier, with mean KL to the oracle family distribution of $8.4$ for the harmful
Llama-3.1-8B versus $3.2$ for Claude Haiku 4.5, tracking the harm rate (Table~\ref{tab:detfilter});
we read this as a diagnostic of directional mismatch, not a direct estimate of $\beta$. Second, a
representative trajectory (App.~\ref{app:cases}): after finding a strong gradient-boosted
configuration in round one, the critic verbally endorses it every subsequent round yet proposes
steadily worse alternatives---Assumption~\ref{ass:waste} realized by a real critic. Because the loop
retains the incumbent, their cost is the forgone exploration unguided search would have reached.

\subsection{A net-benefit criterion from effective search horizon}
\label{subsec:esscriterion}

The following model is deliberately idealized (i.i.d.\ scores, independent round types); its purpose
is to explain the observed qualitative asymmetry and derive intervention targets, not to predict the
sign of every dataset-level effect. Model a critic by two per-round rates: with probability $q$ it is
\emph{competent}, proposing the best of an effective $m$-configuration lookahead ($m\ge1$, a proxy
for how sharply it concentrates on good regions); with probability $\beta$ it is \emph{adversarial}
(Assumption~\ref{ass:waste}); otherwise it selects uniformly. The no-critic loop is the all-uniform
special case ($q{=}\beta{=}0$).

\begin{lemma}[Conditional effective sample size]
\label{lem:ess}
Conditioned on $K_c$ competent and $K_u$ uniform rounds, the final incumbent has the same
distribution as the maximum of $N_{\mathrm{eff}}=mK_c+K_u$ i.i.d.\ score draws.
\end{lemma}
\noindent The proof (App.~\ref{app:proofs}) deletes adversarial rounds (they never enter the
maximum) and treats each competent round as $m$ exchangeable uniform draws.

Lemma~\ref{lem:ess} is exact. Taking expectations,
$\mathbb{E}[N_{\mathrm{eff}}]=T\big(1+q(m-1)-\beta\big)=T_{\mathrm{eff}}$, so $T_{\mathrm{eff}}>T$
exactly when $q(m-1)>\beta$. A per-round comparison with random search makes the role of this
condition precise:

\begin{proposition}[Effective-horizon criterion: necessary condition and tail boundary]
\label{prop:ess}
Let $F$ be the score distribution of a uniform draw and $\varepsilon(b)=1-F(b)$ the probability
that a uniform round improves on incumbent $b$. Under the model above, a critic round improves on
$b$ with higher probability than a uniform round if and only if
$\beta<q\sum_{j=1}^{m-1}F(b)^j$. Consequently the critic can be net-beneficial
($\mathbb{E}[b_T^{\text{critic}}]>\mathbb{E}[b_T^{\text{no-critic}}]$) only if
\[
  \boxed{\,q\,(m-1) \;>\; \beta\,,}
\]
equivalently $T_{\mathrm{eff}}>T$, and in the rare-improvement regime the per-round advantage is
$\varepsilon\,[\,q(m{-}1)-\beta\,]+O(\varepsilon^2)$ as $\varepsilon\to0$.
\end{proposition}
\noindent The condition is necessary, not sufficient: away from the tail, the higher-order terms
matter. With Bernoulli$(1/2)$ scores and $q{=}0.3$, $m{=}3$, $\beta{=}0.3$, $q(m{-}1)=0.6>\beta$, yet a
critic round reaches the top score with probability $0.4625<0.5$, so the critic is worse at every
budget $T$ (App.~\ref{app:proofs}). In the idealized Monte-Carlo simulation of
App.~\ref{app:proofs} at $T{=}8$, the sign of $q(m{-}1)-\beta$ matches the sign of net benefit on
$43/48$ grid cells.

The criterion organizes both axes: capability enters through $q,m$, error structure through
$\beta$. Consistent with this, the family-selection divergence of \S\ref{subsec:mechanism} falls
with critic tier, tracking the harm rate---though we do not directly identify $(q,m,\beta)$ for real
critics, treating the family KL only as a proxy for directional mismatch. Its single actionable
variable, $\beta$, motivates the gates of \S\ref{sec:prevent}, which we evaluate end to end;
because gating also changes exploration, those experiments do not manipulate $\beta$ in isolation.
It also unifies the folklore heuristics as single-variable moves on one
condition---stronger critic raises $q,m$; verification lowers $\beta$; breadth raises $m$
(App.~\ref{app:proofs}).

The criterion does not include search headroom or the frontier--budget setting of
\S\ref{sec:detect}, both of which modulate how strongly directional guidance appears relative to the
no-critic floor (\textsc{credit-g} illustrates the gap: a positive delta with a $\beta$-proxy
indistinguishable from the harmed datasets; further scope discussion in \S\ref{sec:limitations}).

\label{subsec:breadth}%
The criterion also predicts that search \emph{breadth} is a partial structural defense---parallel
$k$-wide search dilutes any single adversarial pick and falls below floor on only $1/12$ datasets
versus $9/12$ for sequential, tree-search, and multi-agent backbones (corollary and derivation in
App.~\ref{app:proofs}; Table~\ref{tab:strategies} in App.~\ref{app:additional}).

\begin{table}[t]
\centering
\caption{Cross-domain test of the mismatch hypothesis (same 8B critic,
four-arm protocol, 8 seeds; No-Critic $\Delta$, 95\% bootstrap CI). Credit
scoring is an \emph{in-domain control} (same domain, same $96$-config space)
yet neutral: only the original tabular pipeline-search setting is harmed.}
\label{tab:crossdomain}
\small
\begin{tabular}{lrrl}
\toprule
Setting (space size) & $\Delta$ & 95\% CI & Verdict \\
\midrule
Tabular AutoML (96)        & $-0.245$ & $[-0.27, -0.22]$   & harm \\
Credit scoring (96)        & $+0.000$ & $[-0.001, +0.002]$ & neutral \\
20NG text-pipe (36)        & $-0.003$ & $[-0.008, +0.001]$ & neutral \\
20NG weak$\to$strong (50)  & $+0.003$ & $[-0.005, +0.010]$ & neutral \\
M4 forecasting (13)        & $+0.001$ & $[+0.001, +0.002]$ & helps \\
RAG retrieval (54)         & $+0.007$ & $[+0.006, +0.008]$ & helps \\
\bottomrule
\end{tabular}
\end{table}

\subsection{A cross-domain test of the prior--reality mismatch hypothesis}
\label{subsec:beyond}

If prior--task mismatch drives directionally adversarial guidance, then, holding the critic fixed,
harm should track where its inductive preferences conflict with the task's performance landscape. We
test this with the same 8B critic and four-arm protocol (8 seeds) across five further settings---credit
scoring, text pipelines, forecasting, retrieval (Table~\ref{tab:crossdomain}). Only the original
\textsc{adult} pipeline-search is clearly negative ($-0.245$); the in-domain credit-scoring control
and both text settings are neutral, and forecasting and retrieval are slightly positive. Holding
the critic fixed, model size cannot explain this variation; the pattern is consistent with the
mismatch hypothesis, which we do not test directly (e.g., by anonymizing configuration names).

\paragraph{Broader implication.}
The analysis suggests a hypothesis for other systems that retain a best-so-far candidate or use
feedback to accept, reject, or redirect future proposals: non-directional evaluator noise may be
substantially less harmful than systematic bias. Extending the quantitative boundary
$q(m{-}1)>\beta$ to self-refinement, verifier-guided decoding, or preference-based optimization
requires setting-specific assumptions and remains future work.

\section{\emph{How}: Verification, Filtering, and Selective Abstention}
\label{sec:prevent}

\subsection{Motivation: the harm appears in a published system and survives prompting}
\label{subsec:motivation}
\label{sec:harm}%
\label{subsec:caafe}%
\label{subsec:prompting}%

Before asking how to prevent harm, we test two natural objections: that it is
specific to our cached frontier, and that it can be prompted away.

\paragraph{The failure reproduces in a published system.}
Running the published \textsc{CAAFE} system \cite{hollmann2024caafe} with the same Llama-3.1-8B on
the eight frozen splits, the LLM-guided system is negative on three datasets (\textsc{adult}
$-0.030$, \textsc{bank-marketing} $-0.054$, \textsc{credit-g} $-0.090$), mean $-0.021$, no gain
above $+0.004$. Because CAAFE differs in feature-engineering space, dynamics, and headroom, we read
this as evidence that below-floor harm is not specific to our cached frontier, not as a per-dataset
replication. A stress test on two public regression tasks likewise goes negative, so the mechanism
is not specific to classification or ROC-AUC (App.~\ref{app:additional}).

\paragraph{Prompting and added compute do not reliably remove it.}
Chain-of-thought and self-consistency shrink the harm but do not reliably flip its sign: on
\textsc{adult}, chain-of-thought lifts the delta from $-0.278$ to $-0.041$ yet three of four
datasets stay below floor, and five-sample self-consistency spends $5\times$ the generations while
doing \emph{worse} than plain chain-of-thought (Table~\ref{tab:family}). Compute does not
monotonically buy safety---the zero-call no-critic condition beats naive, chain-of-thought, and
$40$-generation variants alike (Fig.~\ref{fig:cost}). This sets the prevention bar: remove the loss
without extra critic generations, by targeting the directional-error mechanism of
\S\ref{sec:theory} directly.

\subsection{Prevention objective and evaluation protocol}
\label{subsec:preventsetup}

The prevention objective is deliberately asymmetric. Because the attainable upside on these
frontiers is small (\S\ref{sec:detect}), directional error is costly (\S\ref{sec:theory}), and
added reasoning does not remove the loss (\S\ref{subsec:motivation}), we prioritize
non-inferiority to the no-critic floor over extracting small gains from every critic proposal.
Filtering cannot create search headroom: on a frontier with little headroom, successful prevention
may return the system to the floor without measurable positive gain.

Both methods in this section instantiate one gated policy. At each round a binary gate $g_t$
decides whether the newly generated critique may condition the next proposal:
\begin{equation}
  a_{t+1} \;=\;
  \begin{cases}
    \pi(H_t,\, c_t), & g_t = 1,\\[2pt]
    \pi_0(H_t), & g_t = 0,
  \end{cases}
  \label{eq:gatedpolicy}
\end{equation}
where $\pi$ is the proposer of \S\ref{subsec:loop} and $\pi_0$ is a critique-free fallback policy.
The gate $g_t$ is produced by a training-free validation rule (\textbf{VGC},
\S\ref{subsec:vgc}) or by a learned detector (\S\ref{subsec:detector}). When the validation anchor
is unstable, the policy may abstain from further critic-conditioned decisions and use the uniform
fallback. This prevents additional directional interventions after the switch, but it does not
recover evaluations already consumed by the critic-guided prefix.
The two gates trade data for immediacy: the \textbf{learned detector}
(\S\ref{subsec:detector}) applies when historical labeled runs are available to supervise it
offline; \textbf{VGC} (\S\ref{subsec:vgc}) needs no training data and gates online from the current
run's validation signal---the cold-start default. Low-stability abstention is an empirical fallback,
not a construction-level guarantee relative to full-budget random search.

These methods are motivated by $q(m{-}1)>\beta$, but we do not claim they manipulate $\beta$ in
isolation: gating also changes the history the proposer conditions on and the effective exploration
policy, so the experiments evaluate the entire resulting policy end to end. We report three
dimensions: \emph{safety} (No-Critic delta of the final incumbent), \emph{retained utility}
(relative to the raw critic), and \emph{cost} (configuration evaluations and critic generations); a
policy has \emph{no observed below-floor effect} when $\Delta_{\mathrm{NC}}(D)\ge0$, an empirical
description rather than a safety guarantee. The end-to-end evaluations (\S\ref{subsec:endtoend})
use $T{=}8$, eight paired seeds, and a uniform reject branch.

\subsection{Training-free verification-gated critique}
\label{subsec:vgc}

VGC sets $g_t$ from the most recent realized transition. The proposal $a_t=\pi(H_{t-1},c_{t-1})$ is
evaluated once (counted against the budget); VGC sets $g_t=1$ if $s_{\mathrm{val}}(a_t)$ is within a
tolerance $\tau_t$ of the incumbent's validation utility, and $g_t=0$ otherwise, discarding $c_t$ so
round $t{+}1$ follows $\pi_0$. This is a transition-level trust update: it stops a poorly performing
feedback-conditioned trajectory from propagating, rather than certifying an individual critique's
correctness before use.
Rejected rounds consume budget like any
other round; the gate never triggers an additional same-round evaluation.
configurations and the search history; it never accesses test utilities or the validation
utilities of untried configurations.

\subsection{Reliability-aware abstention}
\label{subsec:abstain}

The gate overrules the critic using the validation anchor, so it should defer when that anchor is
itself unstable. We measure anchor stability by $\widehat{\rho}_t$, the bootstrap stability of the
validation-induced ranking over the configurations tried so far: validation examples are
resampled, each tried configuration's utility is recomputed on the resample, and
$\widehat{\rho}_t$ summarizes the rank agreement across replicates. Computing $\widehat{\rho}_t$
requires the per-example validation predictions of tried configurations; it uses neither test
labels nor the scores of untried configurations
observable proxy for whether the anchor can support gating, not as a calibrated estimate of
test-set reliability. The tolerance interpolates
\begin{equation}
  \tau_t \;=\; \tau_{\min} \;+\; (1-\widehat{\rho}_t)\,(\tau_{\max}-\tau_{\min}),
  \label{eq:margin}
\end{equation}
so that when the ranking is stable the gate is strict ($\tau_t\!\to\!\tau_{\min}$) and when
stability is low the system becomes less willing to overrule the critic with an unreliable signal.
Because a widened tolerance still gates with an untrustworthy signal, a low-stability policy can
abstain and use uniform proposals for the remaining rounds. This limits further exposure but does
not make the complete trajectory distributionally equivalent to no-critic search.

\subsection{A transferable harmful-feedback detector and \textsc{CritFault}}
\label{subsec:detector}

With historical trajectories available, controlled intervention outcomes can supervise a learned
gate. We package the fault-injection trajectories as \textsc{CritFault} ($\sim$3{,}840 labeled
loop transitions across 12 datasets; labels, features, and a separate real-harm relabeling detailed
in App.~\ref{app:critfault}), to our knowledge the first benchmark for detecting harmful critic
feedback in tabular AutoML. The deployable detector reads only evaluated validation utility,
incumbent-relative change, and round fraction. The two structural indicators (untried-utility rank
and neighborhood membership) are retained only for offline diagnosis. Like VGC it produces
$g_t$ after evaluation and before the critique conditions the next proposal
(Eq.~\eqref{eq:gatedpolicy}); unlike post-hoc failure monitors
\cite{pathak2025silentfailures,zhang2025failureattribution}, it prevents propagation online at no
added LLM cost (a gradient-boosted model over the three deployable features).

With all five features, the leave-one-dataset-out AUC is $0.95$ over the 12-dataset fault-injection benchmark and transfers
across critic models (leave-one-model-out mean $0.988$). On six datasets and 384 realized critic
transitions with the same labels and folds, the fully deployable f0--f2 detector attains $0.982$;
adding the two offline structural features raises this to $0.992$
(App.~\ref{app:additional}).
Classification AUC alone, however, does not establish a safe filter; the end-to-end test follows.

\subsection{End-to-end prevention results}
\label{subsec:endtoend}

\paragraph{Paired VGC audit ($T{=}8$).}
We reran fixed and reliability-adaptive tolerances on eight datasets and eight paired seeds, using
a uniform rather than validation-greedy reject branch. Adaptive minus fixed is $-0.00006$ AUC with
a paired 95\% CI of $[-0.00109,0.00090]$; five of eight point estimates are positive. Relative to
the uniform reference, fixed minus uniform is $-0.00032$ $[-0.00140,+0.00072]$ and adaptive minus
uniform is $-0.00038$ $[-0.00166,+0.00078]$; both lower bounds exceed $-0.002$, which supports
exploratory average non-inferiority at a $0.002$ margin, not dataset-wise safety. Thus the data do
not support adaptive superiority or a suite-wide safety guarantee. We retain VGC as an
exploratory gate and treat the fixed observable rule as the primary training-free baseline.

\paragraph{Detector and simple gate online.}
In a deployable end-to-end rerun, both gates use only evaluated utility, incumbent-relative change,
and round fraction; a trigger sends the next proposal to uniform search. Across six datasets, the
raw policy averages $-0.00541$ AUC relative to uniform, the simple non-improvement gate
$+0.00046$, and the learned detector $-0.00015$. The simple gate significantly improves over raw
on bank-marketing and kc1; the detector significantly improves over raw on bank-marketing and
electricity. Neither learned detection nor classification AUC alone establishes floor preservation.

\paragraph{Summary.}
Observable gating reduces the average loss in this suite, with the simple validation rule the most
stable end-to-end baseline. Adaptive tolerance and learned detection can help individual datasets,
but neither supports a universal non-inferiority claim. The operating thresholds
reported here should be recalibrated when the critic, search space, or data regime changes.

\section{Limitations and Ethical Considerations}
\label{sec:limitations}

\textbf{Scope.} The primary estimates concern finite-budget tabular search on small, fixed
frontiers (96 configurations, $T{=}8$), where in-space headroom is below $0.01$ AUC on every
dataset. The below-floor effect shrinks when frontier and horizon grow together
(\S\ref{subsec:attenuate}), and we do not claim harm at long horizons; Proposition~\ref{prop:threshold}
predicts that directional corruption is tolerated once the budget is large relative to
$T_{\mathrm{sat}}$. The capability ladder uses six datasets chosen partly by the 8B critic's
outcome, and per-dataset intervals quantify search randomness conditional on one frozen split.

\textbf{What the audit isolates.} In the reported experiments, critique and proposal share one LLM
call, and the random-feedback arm is an independently seeded random search
(\S\ref{sec:protocol}). The audit therefore measures the complete feedback-conditioned selection
policy, including its parser and fallback, rather than the content of the critique in isolation.
The condition $q(m{-}1)>\beta$ is derived in an idealized model; $(q,m,\beta)$ are not identified
for real critics.

\textbf{Broader impact.} Tabular ML is the production default in high-stakes domains---credit
scoring \cite{lessmann2015benchmarking,baesens2003benchmarking}, fraud detection
\cite{dalpozzolo2014fraud}, clinical risk \cite{christodoulou2019clinical}. In such
threshold-driven applications, even a $\sim\!0.03$ AUC drop can shift a non-trivial slice of the
accept/reject boundary. The work is defensive---high-stakes critics still need human oversight; all
data are public OpenML benchmarks.

\section{Conclusion}
\label{sec:conclusion}

Adding an LLM critic to an iterative search loop is not guaranteed to be harmless. We formalized
\emph{critic safety} and characterized when adding a critic helps or hurts on finite-budget
tabular search: \textbf{when}---harm attenuates with critic capability and varies with frontier and
budget, without a universal capability threshold; \textbf{why}---best-so-far search tolerates
non-directional errors but not directional ones, and an idealized model gives the necessary
condition $q(m{-}1)>\beta$; and \textbf{how}---observable gates that target $\beta$ reduce the
average loss in deployable reruns, without guaranteeing the no-critic baseline on every dataset.
Cross-domain audits are consistent with the hypothesis that harm tracks mismatch between the
critic's priors and the task. The practical takeaway: audit a critic against a no-critic control at
the intended budget before trusting it, and gate or omit it when capability or verification is
insufficient. We will release \textsc{CritFault} and all artifacts upon publication.

\bibliographystyle{plainnat-keepcase}
\bibliography{refs}

\clearpage
\appendix

\section{Metric Definitions}
\label{app:metrics}
For dataset $D$, critic model $M$, and search policy $\Pi$, let $\mathrm{AUC}_\Pi(D,M)$ be the final
best-so-far out-of-sample ROC-AUC (mean over seeds). We report:
\begin{align}
\text{No-Critic delta:}\quad & \Delta(D,M) = \mathrm{AUC}_{\text{real}}(D,M) - \mathrm{AUC}_{\text{no-critic}}(D), \\
\text{Second-replicate delta:}\quad & \Delta_{\text{RF}}(D,M) = \mathrm{AUC}_{\text{real}}(D,M) - \mathrm{AUC}_{\text{rand-fb}}(D), \\
\text{Harm rate:}\quad & \mathrm{HR}(M) = \tfrac{1}{|\mathcal{D}|}\textstyle\sum_{D} \mathbb{1}[\Delta(D,M)<0 \wedge \mathrm{CI\ excl.\ }0], \\
\text{Guard gain:}\quad & G(D,M) = \mathrm{AUC}_{\text{vgc}}(D,M) - \mathrm{AUC}_{\text{real}}(D,M), \\
\text{Resilience thr.:}\quad & p^\star(D) = \min\{p : \mathrm{AUC}(p{=}0)-\mathrm{AUC}(p) > 0.01\}.
\end{align}
In the reported runs the random-feedback arm is an independently seeded random search, so
$\Delta_{\text{RF}}$ is a second-seed version of $\Delta$ rather than a semantic placebo. All CIs are
$10{,}000$-sample nonparametric bootstrap over seeds; paired significance is Wilcoxon
signed-rank across seeds (per dataset) or across datasets (suite level).

\section{Proof Details}
\label{app:proofs}
Statements of Proposition~\ref{prop:random}, Proposition~\ref{prop:threshold},
Lemma~\ref{lem:ess}, and Proposition~\ref{prop:ess} are in \S\ref{sec:theory}; the full proofs
follow.

\paragraph{Proof of Proposition~\ref{prop:random} (uniform corruption preserves the floor).}
Selection without replacement of $T$ configurations uniformly at random induces the same
distribution over the \emph{set} $S_T=\{\pi(1),\dots,\pi(T)\}$ regardless of the order or of which
rounds were labeled ``corrupted,'' because each round's draw is uniform over the remaining
configurations in both the corrupted and uncorrupted case. Since $b_T=\max_{a\in S_T} s(a)$ depends
only on the set $S_T$, its distribution, and hence its expectation, is identical to the no-critic
loop, which draws $S_T$ the same way. The claim holds for all $p$ because the mixing weight between
two identical sampling processes is irrelevant. The proposition requires only
sampling-without-replacement exchangeability and that $b_T$ be a function of the visited set; it
makes no assumption on $s(\cdot)$. As stated in \S\ref{subsec:randomfree}, the hypothesis is that
selection is uniform in \emph{every} round; the result is a floor-preservation statement for an
uninformative policy, not a claim that randomly corrupting an informative critic is costless.

\paragraph{Proof of Proposition~\ref{prop:threshold} (resilience under adversarial corruption).}
By Eq.~\eqref{eq:mono} the incumbent is non-decreasing, and by Assumption~\ref{ass:waste} corrupted
rounds leave it unchanged; hence $b_T$ equals the incumbent obtained from the subsequence of
effective rounds alone. Let $K\sim\mathrm{Binomial}(T,1-p)$ denote the (random) number of effective
rounds, and let $E=\{K\ge T_{\mathrm{sat}}\}$. On $E$, the definition of the saturation budget
gives $\mathbb{E}[\,b^\star-b_T \mid E\,]\le\delta$. On the complement $E^{c}$, scores lie in
$[0,1]$, so $b^\star-b_T\le 1$ pointwise. Splitting the expectation,
\begin{align*}
  \mathbb{E}[\,b^\star-b_T\,]
  &=
  \mathbb{E}[\,b^\star-b_T \mid E\,]\Pr[E] \;+\; \mathbb{E}[\,b^\star-b_T \mid E^{c}\,]\Pr[E^{c}]\\
  &\le
  \delta \;+\; \Pr\!\big[K<T_{\mathrm{sat}}\big],
\end{align*}
which is the stated bound. For a tolerated additional failure probability $\epsilon$, the loop is
$(\delta+\epsilon)$-robust whenever $\Pr[\mathrm{Binomial}(T,1-p)<T_{\mathrm{sat}}]\le\epsilon$;
since this tail probability is nondecreasing in $p$, the threshold
$p^\star(\epsilon)=\sup\{p:\Pr[\mathrm{Binomial}(T,1-p)<T_{\mathrm{sat}}]\le\epsilon\}$ is
well-defined. The expectation-level approximation $p^\star\approx 1-T_{\mathrm{sat}}/T$ arises from
replacing $K$ by its mean $(1-p)T$ and requiring $(1-p)T\ge T_{\mathrm{sat}}$; at the paper's scale
($T{=}8$, $T_{\mathrm{sat}}\in[1,3]$) it predicts $p^\star\in[0.63,0.88]$, and the binomial tail at
these parameters is small enough that the probabilistic and expectation-level thresholds agree to
the granularity of the swept grid $\mathcal{P}$. Two remarks on the hypotheses.
Assumption~\ref{ass:waste} (adversarial rounds do not raise the incumbent) holds once $b_{t-1}$
exceeds the low-score region, empirically after round~1 on all nine fault-injection datasets, and the
$90.5\%$ empirical satisfaction rate reported in \S\ref{subsec:adversarial} means the deleted-round
argument applies to the large majority of adversarial rounds, with the bound degrading gracefully
in the remainder. The event $E$ conditions on the count $K$ only, not on which rounds are
effective; this is licensed by the exchangeability of uniform-without-replacement selection over
the effective subsequence.

\paragraph{Proof of Lemma~\ref{lem:ess} (conditional effective sample size).}
By elitist monotonicity (Eq.~\eqref{eq:mono}) the incumbent depends only on the multiset of
realized values, and adversarial rounds (Assumption~\ref{ass:waste}) leave it unchanged, so they
may be deleted from the multiset without changing $b_T$. Condition on $K_c=k_c$ competent and
$K_u=k_u$ uniform rounds. A uniform round contributes one i.i.d.\ score draw. A competent round
proposes the best of an effective $m$-configuration lookahead, so its realized value is distributed
as the maximum of $m$ i.i.d.\ draws; for the purpose of the \emph{outer} maximum, such a round is
exchangeable with $m$ independent uniform draws (the maximum of maxima over disjoint blocks equals
the maximum of the pooled draws). Hence, conditioned on $(k_c,k_u)$, the final incumbent is
distributed as the maximum of $N_{\mathrm{eff}}=m k_c+k_u$ i.i.d.\ draws, as claimed. The lemma is
exact under the model's assumptions (i.i.d.\ scores, independent round types).

\paragraph{Proof of Proposition~\ref{prop:ess} (necessary condition and tail boundary).}
Taking expectations over round types in Lemma~\ref{lem:ess} gives
$\mathbb{E}[N_{\mathrm{eff}}]=T(1+q(m-1)-\beta)=T_{\mathrm{eff}}$, so $T_{\mathrm{eff}}>T$ if and only if
$q(m-1)>\beta$. Under the model, rounds are independent and each uniform round contributes one draw
from $F$, so it suffices to compare the score distribution of a single round. Treating an
adversarial round as contributing no improvement (Assumption~\ref{ass:waste}), the per-round CDF of
a critic round is
\[
  G(x) \;=\; \beta + (1-q-\beta)\,F(x) + q\,F(x)^m .
\]
For an incumbent $b$, a uniform round improves with probability $1-F(b)$ and a critic round with
probability $1-G(b)$. Their difference is
\[
  F(b)-G(b) \;=\; q\big(F(b)-F(b)^m\big)-\beta\big(1-F(b)\big)
  \;=\; \big(1-F(b)\big)\Big[\,q\sum_{j=1}^{m-1}F(b)^j-\beta\,\Big],
\]
which is positive exactly when $\beta<q\sum_{j=1}^{m-1}F(b)^j$. Because $\sum_{j=1}^{m-1}F(b)^j<m-1$
whenever $F(b)<1$, the condition $q(m-1)\le\beta$ implies $G\ge F$ everywhere: each critic round is
stochastically dominated by a uniform round, and so is the final incumbent, giving
$\mathbb{E}[b_T^{\text{critic}}]\le\mathbb{E}[b_T^{\text{no-critic}}]$ for every $T$. Hence
$q(m-1)>\beta$ is necessary for net benefit. Writing $\varepsilon=1-F(b)$ and expanding
$F(b)^j=(1-\varepsilon)^j$ gives $F(b)-G(b)=\varepsilon\,[\,q(m-1)-\beta\,]+O(\varepsilon^2)$, so
$q(m-1)>\beta$ is the leading-order boundary in the rare-improvement regime $\varepsilon\to0$.

The condition is not sufficient. Take Bernoulli$(1/2)$ scores, initial incumbent $0$, and
$q{=}0.3$, $m{=}3$, $\beta{=}0.3$, so $q(m-1)=0.6>\beta$. A critic round reaches score $1$ with
probability $0.3\,(1-2^{-3})+0.4\cdot\tfrac12=0.4625$, versus $0.5$ for a uniform round, so
$\mathbb{E}[b_T^{\text{critic}}]=1-0.5375^T<1-0.5^T=\mathbb{E}[b_T^{\text{no-critic}}]$ for every
$T\ge1$. Here $\varepsilon=0.5$ is far from the tail and the higher-order terms dominate; for this
distribution the actual boundary is $\beta=0.225$ at every $T$. Expected final utility therefore
depends on the full score distribution, and we use $q(m-1)>\beta$ only as a necessary condition and
tail approximation.

\paragraph{The criterion subsumes existing critic heuristics.} The folklore practices for ``when to
trust an LLM critic'' are each a single-variable move on the same condition $q(m{-}1)>\beta$:
\emph{use a stronger critic} raises $q,m$; \emph{verify before trusting} (self-consistency,
LLM-as-judge gating) lowers $\beta$; \emph{search broadly} (parallel/tree policies) raises the
effective lookahead $m$ (corollary below). This is why none of them is universally safe: raising
capability fails if $\beta$ stays high (a fluent but confidently wrong critic), and verifying fails
if the verifier is itself unreliable (\S\ref{sec:prevent}, low-$\widehat{\rho}$). The condition
indicates which move is needed when; our prevention (\S\ref{sec:prevent}) targets $\beta$.

\paragraph{Corollary: breadth is a structural defense.} Proposition~\ref{prop:ess} suggests
that \emph{parallel} search resists harm. Expanding $k$ candidates per round and keeping the best
raises the effective lookahead ($m\to m'\!\ge\!m$), which widens the range of adversarial rates
for which net benefit is possible to $\beta<q(m'{-}1)$; a $k$-wide round also dilutes a single adversarial
pick to weight $1/k$. The boundary thus shifts toward net-benefit as breadth grows, predicting that
breadth-oriented policies are hard to harm, matching the measured backbones
(Table~\ref{tab:strategies}: parallel $k$-wide search below floor on $1/12$ datasets versus $9/12$
for sequential, tree-search, and multi-agent; \S\ref{subsec:breadth}).

\paragraph{Monte-Carlo check of the criterion.} We cannot check $q(m{-}1)>\beta$ per real
dataset (sequential critics violate the independent-round assumption), so we simulate the idealized
loop. For $q{=}0.3$, $m{=}3$ ($q(m{-}1)=0.60$), the measured sign-flip boundary in our simulation is
$0.42$, $0.53$, $0.56$, and $0.60$ at $T{=}8,32,128,512$, and at $T{=}8$ the sign of
$q(m{-}1)-\beta$ matches the sign of net benefit on $43/48$ grid cells. This agreement is specific
to the simulated score distribution: the Bernoulli example above has boundary $0.225$ at every
$T$, so the simulation illustrates the tail approximation rather than a general convergence
result.

\section{Implementation Details}
\label{app:impl}

\begin{algorithm}[htbp]
\caption{Actor--Critic AutoML loop}
\label{alg:loop}
\small
\begin{algorithmic}[1]
\State \textbf{Input:} $D_\text{tr}, D_\text{val}$, rounds $T$, patience $p$
\State $f^\ast \gets \textsc{Train}(D_\text{tr},D_\text{val})$;\; $r^\ast \gets \textsc{Metric}(f^\ast)$;\; $c \gets 0$
\For{$t = 1$ to $T$}
  \State $a_t \gets \textsc{Actor}(s_t, h_{t-1}, \text{corr}_{t-1})$ \Comment{propose}
  \State $f_t \gets \textsc{Pipeline}(D_\text{tr}, D_\text{val}; a_t)$ \Comment{execute}
  \State $r_t \gets \textsc{Metric}(f_t)$
  \State $\text{corr}_t \gets \textsc{Critic}(a_t, r_t, h_{t-1})$ \Comment{critique}
  \State $h_t \gets h_{t-1} \cup \{(a_t, r_t, \text{corr}_t)\}$
  \If{$r_t > r^\ast$} $f^\ast \gets f_t$;\; $r^\ast \gets r_t$;\; $c \gets 0$
  \Else\ $c \gets c{+}1$;\ \textbf{if } $c \ge p$ \textbf{ break}
  \EndIf
\EndFor
\State \Return $f^\ast$
\end{algorithmic}
\end{algorithm}

In the reported experiments, \textsc{Actor} and \textsc{Critic} are realized by a single LLM call
per round that reads the evaluated history and the menu of untried configurations and names the
next configuration; a response that names no untried configuration triggers a uniform fallback
(\S\ref{subsec:frontier}).

Search space: 8 model families $\times$ 4 transform sets $\times$ 3 hyperparameter presets ($=96$).
The four feature-transformation sets are generated by applying the same fixed
construction rule separately to each dataset, using its two highest-variance
numeric features as transformation anchors; the complete configuration list is
released with the code.
Loop: $T{=}8$ rounds (legacy VGC runs: $T{=}10$), greedy incumbent, no explicit patience. LLM critics served via
Amazon Bedrock Converse, temperature $0$, $\le 1536$ max tokens; models are the published open-weights
Llama-3.1-8B / Llama-3.3-70B inference profiles and Claude Haiku 4.5. All tree models pinned
single-threaded (\texttt{OMP\_NUM\_THREADS=1}) to avoid oversubscription. Detector: gradient-boosted
trees, 100 estimators, depth 3, evaluated leave-one-dataset-out.

\subsection{Formal setup details (moved from \S\ref{sec:protocol})}
\label{app:formal-setup}

\paragraph{History and feedback maps.}
Let $H_{t-1}=\{(a_i,s_{\mathrm{val}}(a_i))\}_{i=1}^{t-1}$ denote the evaluated
configurations and their observed validation utilities, and $c_{t-1}$ the most
recent feedback. At round $t$ the proposer selects
$a_t=\pi(H_{t-1},c_{t-1})$; after evaluation and the incumbent update
$\hat a_t=\arg\max_{a\in\{a_1,\ldots,a_t\}} s_{\mathrm{val}}(a)$, the critic
produces $c_t=C(a_t,s_{\mathrm{val}}(a_t),H_{t-1})$, which may influence the
next proposal. In the no-critic condition $c_t=\varnothing$ reduces selection
to uniform exploration over untried configurations; the oracle bypasses the
language channel entirely and upper-bounds any feedback-mediated policy on the
same frontier.

\paragraph{Terminology.}
Because the critic only shapes which configurations are explored---it holds no
value function and receives no reward---we refer to the system as a
\emph{proposer--critic loop} rather than an actor--critic method with a
reinforcement-learning interpretation. The loop is externally grounded (the
critic sees held-out validation utilities), unlike intrinsic self-critique in
which a model evaluates its own output without an external task signal
\cite{shinn2023reflexion}; the validation signal both makes critic guidance
auditable and supplies the evidence used to gate harmful feedback in
\S\ref{sec:prevent}.

\paragraph{Cached-frontier access rules.}
Before any search condition runs, every configuration in $\mathcal{A}$ is
evaluated once on the frozen split and its validation and test utilities
cached. During a run, the search may read $s_{\mathrm{val}}(a)$ only after
selecting $a$; test utilities remain hidden until the final incumbent is
fixed. Observed differences between arms therefore cannot be attributed to
pipeline-training randomness, execution failures, resampling, or access to
different candidate pools.

\paragraph{Resilience threshold for fault injection.}
For dataset $D$, corruption mode $c$, and rate $p$, let $J_{D,c}(p)$ denote
the expected test utility of the final validation incumbent. The
mode-specific resilience threshold is the first evaluated corruption rate at
which degradation exceeds a fixed tolerance,
\[
p_{D,c}^{\star}
=
\min\left\{p\in\mathcal{P}:\, J_{D,c}(0)-J_{D,c}(p)>\delta_{\mathrm{res}}\right\},
\]
with $\delta_{\mathrm{res}}=0.01$~AUC,
where $\mathcal{P}\subset[0,1]$ is the swept grid. Two configurations are
neighbors (for \textsc{Mislead}) when they differ in exactly one pipeline
component. The tolerance $\delta_{\mathrm{res}}$ defines meaningful
degradation under corruption.

\paragraph{Statistical assessment of below-floor prevalence.}
Evidence for the below-floor direction is assessed at two levels: per-dataset
bootstrap confidence intervals quantify task-level uncertainty, while an
exact suite-level sign-flip test over all $2^8$ sign assignments evaluates
whether negative
$\Delta_{\mathrm{NC}}$ occur systematically across datasets. Prevalence
statements such as ``below floor on 7 of 8 datasets'' refer to the signs of
dataset-level point estimates; the headline inferential claim relies on the
suite-level test rather than requiring every individual interval to exclude
zero.

\paragraph{Analysis-specific dataset subsets.}
The primary real-critic audit uses the eight binary-classification datasets.
Controlled fault injection uses the nine datasets other than the three
near-zero-headroom datasets. The capability ladder is a targeted follow-up on
six binary datasets selected partly on the reference 8B critic's outcome; it is
therefore used only to study how an established effect changes with
capability, not to estimate prevalence of harm. The \textsc{CritFault}
detector is evaluated on the full 12-dataset suite, including the three
near-zero-headroom datasets, where harmful transitions are less sharply
defined.

\subsection{Analysis-specific experimental settings}
\label{app:exp-settings}

Table~\ref{tab:analysis-scope} summarizes the datasets, search budgets,
seed counts, and decoding settings used in each analysis. The primary
four-arm audit and capability ladder use $T=8$ and eight matched seeds.
Controlled fault injection uses 30 seeds. The end-to-end prevention reruns of
\S\ref{subsec:endtoend} use $T=8$ and eight paired seeds; the legacy
prevention results in App.~\ref{app:additional} use the budgets stated in
their captions. The expanded-frontier experiment uses $T=20$, and the
wide-space stress test uses $T=16$. Unless otherwise stated, critic decoding
uses temperature $0$; stochastic-decoding robustness experiments use
temperature $0.7$.

Critics are accessed through a provider-agnostic Amazon Bedrock Converse
interface. Exact model identifiers, providers, and release generations are
listed with the capability ladder (Table~\ref{tab:ladder6}); all critics use
their default inference profiles with a $1536$-token generation cap. All
prompts, cached frontiers, frozen splits,
configuration definitions, and evaluation scripts will be released with the
paper.

\begin{table}[t]
\centering
\caption{Experimental scope by analysis. Exact dataset names and model
settings are reported in the appendix.}
\label{tab:analysis-scope}
\begin{tabular}{@{}lll@{}}
\toprule
Analysis & Datasets & Purpose \\
\midrule
Real-critic audit & 8 binary & Below-floor harm \\
Fault injection & 9 (see text) & Error structure \\
Capability ladder & 6 targeted & Capability attenuation \\
\textsc{CritFault} & 12 total & Feedback detection \\
\bottomrule
\end{tabular}
\end{table}

\section{Additional Experiments}
\label{app:additional}

\paragraph{Per-dataset detector generalization (leave-one-dataset-out).}
Table~\ref{tab:detector} gives the per-dataset breakdown behind the mean AUC $0.982$ reported in
\S\ref{sec:prevent}: on injection labels with all five features, the bad-feedback signature
transfers to each of the nine datasets outside the near-zero-headroom group (each $\ge 0.96$, mean
$0.982$) and is lower (mean $0.851$) on the three near-zero-headroom datasets, where harmful
transitions are less sharply defined. This $0.982$ is distinct from the $0.982$ of the deployable
f0--f2 detector on real-harm labels (\S\ref{subsec:detector}).

\begin{table}[htbp]
\centering
\caption{Bad-feedback detector, \textbf{leave-one-dataset-out} AUC (train on 11 datasets, test on the
held-out one), injection labels, all five features (including the two offline structural
features). Mean $0.982$ on the nine datasets outside the near-zero-headroom group (every one
$\ge 0.96$) and $0.851$ on the three near-zero-headroom datasets (segment/mfeat/splice), where
nearly every configuration scores near the oracle.}
\label{tab:detector}
\small
\begin{tabular}{lrlr}
\toprule
\multicolumn{4}{l}{\emph{Nine datasets} (mean $0.982$, all $\ge 0.96$)}\\
\midrule
bank-marketing & 0.990 & letter        & 0.983 \\
credit-g       & 0.989 & churn         & 0.980 \\
phoneme        & 0.988 & jm1           & 0.974 \\
adult          & 0.987 & kc1           & 0.966 \\
electricity    & 0.985 &               &       \\
\midrule
\multicolumn{4}{l}{\emph{Near-zero headroom} (mean $0.851$)}\\
\midrule
mfeat-factors  & 0.962 & segment       & 0.732 \\
splice         & 0.860 &               &       \\
\midrule
\multicolumn{4}{c}{\textbf{Overall mean 0.950} (12 datasets, 3{,}840 transitions)} \\
\bottomrule
\end{tabular}
\end{table}

\paragraph{Capability ladder, full per-dataset numbers.} Table~\ref{tab:ladder6-full} gives the complete
seven-tier, four-provider ladder behind Fig.~\ref{fig:axis1}a. The 2024 $8$B critic is negative on
$5/6$ datasets with losses up to $0.245$; every $\ge$14B critic stays within $0.031$ of zero, but
negative point estimates remain (three of six at $14$B, five of six at Llama-3.3-$70$B). The ladder
supports attenuation of the large losses, not a capability threshold.

\begin{table}[htbp]
\centering
\caption{Capability ladder across a \emph{seven-tier, four-provider} span: real-critic No-Critic
$\Delta$ (AUC, 8 seeds, identical frozen splits) on six discriminating datasets (full per-dataset
version of Table~\ref{tab:ladder6}). The 2024-generation $8$B critic (Meta) is negative on $5/6$;
the larger critics are within $0.031$ of zero but not uniformly non-negative (Llama-3.3-$70$B reaches
$-0.031$ on \textsc{electricity} and $-0.023$ on \textsc{kc1}). The pattern is not monotone in size,
provider, or release date.}
\label{tab:ladder6-full}
\small
\setlength{\tabcolsep}{3.5pt}
\begin{tabular}{l@{\hskip 5pt}rrrrrrr}
\toprule
Dataset & 3B$^{25}$ & 8B$^{24}$ & 8B$^{25}$ & 14B & 32B & 70B & H4.5 \\
\midrule
adult          & $-.210$ & $-.245$ & $+.001$ & $+.001$ & $+.001$ & $-.011$ & $+.002$ \\
electricity    & $-.064$ & $-.072$ & $-.012$ & $+.012$ & $+.008$ & $-.031$ & $+.006$ \\
jm1            & $-.008$ & $-.080$ & $-.003$ & $-.002$ & $-.000$ & $-.011$ & $+.002$ \\
kc1            & $+.000$ & $-.028$ & $-.000$ & $-.001$ & $-.001$ & $-.023$ & $-.000$ \\
bank-marketing &         & $-.007$ & $+.000$ & $+.000$ & $+.001$ & $-.007$ & $+.001$ \\
credit-g       &         & $+.006$ & $-.005$ & $-.012$ & $-.007$ & $+.006$ & $-.001$ \\
\bottomrule
\end{tabular}\\[2pt]
\scriptsize 3B$^{25}${=}Ministral-3B (Mistral '25); 8B$^{24}${=}Llama-3.1 (Meta '24);
8B$^{25}${=}Ministral-3-8B (Mistral '25); 14B{=}Ministral-14B; 32B{=}Qwen3; 70B{=}Llama-3.3;
H4.5{=}Haiku\,4.5 (Anthropic).
\end{table}

\paragraph{Real SOTA baseline.} Table~\ref{tab:sota} compares our loop against AutoGluon-Tabular on
identical frozen public splits, confirming our search space is SOTA-competitive (in-space oracle
matches or beats AutoGluon on 5/8 datasets).

\begin{table}[htbp]
\centering
\caption{Competitive AutoML setup. Test ROC-AUC on identical frozen public splits: our no-critic
floor and in-space oracle vs.\ \textbf{AutoGluon-Tabular} \cite{erickson2020autogluon} (real SOTA,
$120$s matched budget, \texttt{medium\_quality}). The in-space oracle matches or beats AutoGluon on
$5$ of $8$ datasets; the harm we measure is thus relative to a SOTA-competitive baseline.}
\label{tab:sota}
\small
\begin{tabular}{lrrr}
\toprule
Dataset & no-critic & oracle & AutoGluon \\
\midrule
electricity    & .9411 & \textbf{.9542} & .9316 \\
credit-g       & .8017 & \textbf{.8085} & .7849 \\
churn          & .9412 & \textbf{.9463} & .9264 \\
adult          & .9287 & \textbf{.9302} & .9236 \\
bank-marketing & .9297 & \textbf{.9328} & .9309 \\
kc1            & .8279 & .8314 & \textbf{.8378} \\
phoneme        & .9433 & .9483 & \textbf{.9549} \\
jm1            & .7290 & .7320 & \textbf{.7484} \\
\bottomrule
\end{tabular}
\end{table}

\paragraph{Space size alone is not protective.} Two observations should be kept apart. \emph{(i)~For
the real $8$B critic}, harm is largest on the $96$-config frontier at $T{=}8$ and largely disappears
on the $250$-config frontier at $T{=}20$ (\S\ref{sec:detect}: \textsc{adult}
$-0.245\!\to\!+0.0003$); because frontier and horizon change together, this does not identify the
cause. \emph{(ii)~For a \emph{modeled} systematically-adversarial critic} (not the real LLM), harm
grows as the space enlarges $2.6\times$ ($96\!\to\!250$ configs) on a wide weak-to-strong space with
real headroom (Table~\ref{tab:spacesize}). Enlarging the space is therefore not by itself a defense
against a directional critic.

\begin{table}[htbp]
\centering
\caption{Enlarging the space does not protect against a modeled systematically-adversarial
critic. Wide weak-to-strong space with real headroom ($1.7$--$5.6\%$ oracle-over-mean AUC); enlarging the space $2.6\times$ ($96\!\to\!250$
configs) with a modeled systematically-adversarial critic (No-Critic $\Delta$, AUC$\times 10^3$, $20$
seeds, $T{=}16$, 95\% bootstrap CI). Harm grows with space size on all five datasets.}
\label{tab:spacesize}
\small
\begin{tabular}{lrrr}
\toprule
Dataset & headroom & $\Delta$ @96 & $\Delta$ @250 \\
\midrule
electricity    & $.056$ & $-30.8$ [$-43.6,-18.9$] & $-36.0$ [$-48.1,-24.4$] \\
credit-g       & $.042$ & $-22.2$ [$-29.0,-15.1$] & $-28.3$ [$-38.8,-19.7$] \\
kc1            & $.030$ & $-14.3$ [$-19.7,-8.9$] & $-25.4$ [$-36.3,-15.9$] \\
jm1            & $.028$ & $-14.9$ [$-19.6,-10.2$] & $-15.9$ [$-24.0,-8.6$] \\
adult          & $.017$ & $-5.8$ [$-9.0,-3.2$] & $-13.2$ [$-18.1,-8.9$] \\
\bottomrule
\end{tabular}
\end{table}

\begin{figure}[htbp]
\centering
\includegraphics[width=0.5\linewidth]{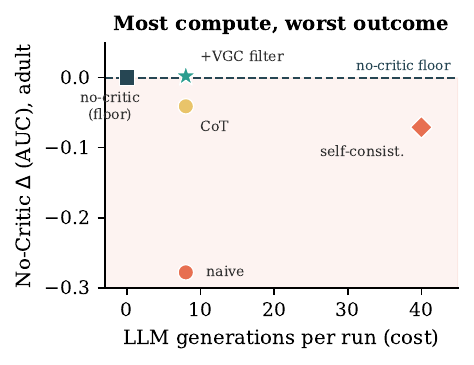}
\caption{Accuracy--cost trade-off for the weak 8B critic (\textsc{adult}). Naive,
chain-of-thought, and $5\times$-cost self-consistency all spend real LLM generations to land
\emph{below} the zero-cost no-critic floor. The ``+VGC filter'' point uses the legacy learned
filter (validation-best reject branch; see above).}
\label{fig:cost}
\end{figure}

\paragraph{Legacy prevention protocol.} Several prevention results in this appendix
(Table~\ref{tab:detfilter}, the last row of Table~\ref{tab:family}, the VGC column of
Table~\ref{tab:finetune}, Fig.~\ref{fig:cost}, and the second-backbone paragraph) come from an
earlier protocol whose reject branch selects the untried configuration with the best cached
validation utility; its learned filter also uses the two offline structural features. Both read
validation utilities of untried configurations, which a deployed loop cannot observe, so we report
these results as exploratory. The deployable reruns in \S\ref{subsec:endtoend} use a uniform reject
branch and only observed features.

\paragraph{Detector-as-online-filter and the harm mechanism (legacy protocol).}
Table~\ref{tab:detfilter} reports the family-selection KL between the critic's picked model-family
distribution and the oracle-optimal one: a harmful raw critic over-commits to weak families (high
KL$_\text{raw}$). Under the legacy protocol above, the filtered policy ends above the no-critic floor
on \textsc{adult}, \textsc{jm1}, and \textsc{kc1} and below it on \textsc{credit-g}, with lower KL.

\begin{table}[htbp]
\centering
\caption{Detector-as-online-filter under the legacy protocol (validation-best reject branch, five
features) and the family-selection KL (Llama-3.1-8B critic, 5 seeds, $T{=}10$, LODO-trained, reject
threshold $\tau{=}0.6$). KL is between
the critic's picked model-family distribution and the oracle-optimal one.}
\label{tab:detfilter}
\small
\begin{tabular}{lrrrr}
\toprule
Data & raw$-$nc & filter$-$nc & KL$_\text{raw}$ & KL$_\text{filter}$ \\
\midrule
adult    & $-.0002$ & $+.0015$ & $8.96$ & $0.15$ \\
jm1      & $-.0028$ & $+.0036$ & $8.70$ & $0.37$ \\
kc1      & $-.0045$ & $+.0018$ & $7.32$ & $5.17$ \\
credit-g & $-.0036$ & $-.0017$ & $8.44$ & $1.65$ \\
\bottomrule
\end{tabular}
\end{table}

\paragraph{Harm generalizes across search backbones (full data).} Reimplementing four search paradigms
over the identical frontier and contrasting no-critic vs.\ a systematically-adversarial critic (30
seeds), the harmful-critic delta is negative for every paradigm (Table~\ref{tab:strategies}), largest
for sequential (which commits hardest) and smallest for breadth-oriented policies---so the harm is a
property of critic-guided search itself, not of any one control flow, and breadth is a partial
structural defense.

\begin{table}[htbp]
\centering
\caption{Harm generalizes across search backbones. Mean harmful-critic $\Delta$ (AUC vs.\ no-critic)
over 12 datasets, 30 seeds; $n_{<0}$ = datasets with $\Delta<-0.002$. A systematically adversarial
critic degrades sequential, tree-search, and multi-agent search alike; parallel $k$-wide expansion is
nearly immune---breadth hedges against an unreliable critic.}
\label{tab:strategies}
\small
\begin{tabular}{lrr}
\toprule
Search policy & mean harmful $\Delta$ & $n_{<0}$/12 \\
\midrule
sequential          & $-0.0183$ & 9 \\
multi-agent (pooled)& $-0.0095$ & 9 \\
tree-search (UCB1)  & $-0.0080$ & 9 \\
parallel $k$-wide   & $-0.0004$ & 1 \\
\bottomrule
\end{tabular}
\end{table}

\paragraph{Prompting family.} Chain-of-thought and self-consistency (\S\ref{sec:harm}) reduce
but do not eliminate harm (Table~\ref{tab:family}): both stay negative on $3/4$ discriminating
datasets, and added compute is not monotone help---self-consistency at $5\times$ the calls scores
$-0.071$ on \textsc{adult}, \emph{worse} than plain CoT's $-0.041$, so the failure is not merely
insufficient compute. The last row uses the legacy learned filter (validation-best reject branch)
and is not a deployable result.

\begin{table}[htbp]
\centering
\caption{Harm is not a naive-prompting artifact. No-Critic $\Delta$ (AUC, 8 seeds, single controlled
run) by critic \emph{method family}, Llama-3.1-8B. Chain-of-thought and self-consistency (\#calls =
LLM generations/run) \emph{reduce} but do not eliminate harm (negative on 3/4; self-consistency worse
than CoT on \textsc{adult} at $5\times$ cost). The last row uses the legacy filter protocol.}
\label{tab:family}
\small
\begin{tabular}{lrrrr}
\toprule
Family (\#calls) & \textsc{adult} & \textsc{jm1} & \textsc{credit-g} & \textsc{kc1} \\
\midrule
no-critic (0)            & $.000$ & $.000$ & $.000$ & $.000$ \\
naive (8)                & $-.278$ & $-.018$ & $+.006$ & $-.019$ \\
CoT (8)                  & $-.041$ & $-.014$ & $+.001$ & $-.006$ \\
self-consistency (40)    & $-.071$ & $+.001$ & $+.003$ & $-.017$ \\
\midrule
+ legacy filter (8)      & $+.002$ & $+.004$ & $-.002$ & $+.002$ \\
\bottomrule
\end{tabular}
\end{table}

\paragraph{Harm is bad advice, not bad formatting (full data).} A natural worry is that the $8$B harm
is merely a format-compliance artifact---a failed parse triggering a random fallback. The data rule
this out: harm and parse-failure are \emph{anti}-correlated across the suite. \textsc{adult}, the most
harmed dataset ($-0.245$), has \emph{zero} fallbacks in $64$ critic rounds, so its harm is entirely
substantive; \textsc{credit-g}, with the \emph{most} fallbacks ($18/64$), is not harmed at all
($+0.006$), because a failed parse degenerates to random search (the floor), never below it. Format
reliability does improve with scale, but the capability effect is driven by proposal \emph{quality}: a
small critic that parses perfectly still steers confidently into low-value regions.

\paragraph{Second backbone (legacy protocol).} On a Llama-3.3-70B critic (8 seeds), legacy VGC
moves \textsc{jm1} from a raw $-0.0010$ to $+0.0036$ and leaves \textsc{adult} essentially unchanged.

\paragraph{Can supervised fine-tuning replace the guard?} A natural objection is to fix the
weak critic by \emph{fine-tuning} it rather than gating it. We test this directly under
\emph{leave-one-dataset-out}: for each dataset we QLoRA-fine-tune Llama-3.1-8B on oracle picks (the
truly-best untried config each round) from the \emph{other seven} datasets, then evaluate the
fine-tuned critic on the held-out eighth---which it never saw in training---in the exact $T{=}8$
setup of \S\ref{sec:detect} (Table~\ref{tab:finetune}, 8 seeds, 95\% bootstrap CI). Fine-tuning does
avert the collapse: on every held-out dataset the No-Critic $\Delta$ returns to within $\pm0.003$
AUC of the floor (e.g.\ \textsc{adult} $-0.245\!\to\!+0.0005$, \textsc{jm1} $-0.080\!\to\!-0.0008$),
with no dataset below floor beyond noise. But it only recovers the \emph{floor}---no held-out
dataset shows a positive delta whose CI excludes zero---and it requires oracle supervision
unavailable at deployment (the oracle label is precisely what the critic cannot see) plus per-suite
GPU training. The VGC column is from the legacy protocol (validation-best reject branch) and is
shown for reference only; it is not a deployable comparison.

\begin{table}[t]
\centering
\caption{Supervised fine-tuning vs.\ VGC under \emph{leave-one-dataset-out} (No-Critic $\Delta$,
AUC$\times10^3$, 8 seeds, 95\% bootstrap CI). For each row the critic is QLoRA-fine-tuned on oracle
picks from the other seven datasets and evaluated on the held-out one. Fine-tuning averts the
collapse on all eight but only returns to the floor (no CI excludes zero positively) and needs oracle
labels. The VGC column uses the legacy protocol.}
\label{tab:finetune}
\small
\begin{tabular}{lrrr}
\toprule
Held-out dataset & raw (\S\ref{sec:detect}) & Fine-tuned (LOO) & VGC \\
\midrule
\textsc{adult}          & $-245$ & $+0.5\ [-0.5,+1.4]$  & $+1.0\ [0.0,+1.9]$ \\
\textsc{electricity}    & $-72$  & $+1.6\ [-2.1,+7.0]$  & $+3.5\ [-1.5,+8.2]$ \\
\textsc{jm1}            & $-80$  & $-0.8\ [-3.4,+1.7]$  & $+1.7\ [-1.7,+4.5]$ \\
\textsc{kc1}            & $-28$  & $-3.1\ [-9.4,+1.2]$  & $-1.2\ [-3.9,+1.9]$ \\
\textsc{churn}          & $-2$   & $+1.7\ [-1.1,+4.5]$  & $-1.8\ [-6.1,+1.9]$ \\
\textsc{bank-marketing} & $-7$   & $-0.7\ [-1.4,+0.0]$  & $-0.7\ [-6.6,+3.6]$ \\
\textsc{phoneme}        & $-9$   & $-0.5\ [-2.0,+0.7]$  & $+1.2\ [-1.0,+3.4]$ \\
\textsc{credit-g}       & $+6$   & $+0.9\ [-5.3,+6.8]$  & $+4.9\ [+0.7,+10.2]$ \\
\bottomrule
\end{tabular}
\end{table}

\paragraph{Harm generalizes to regression.} The main study is binary classification; harm is not
specific to it. On public regression datasets (California housing, diabetes; weak-to-strong regressor
space scored by test $R^2$, 20 seeds), a systematically-adversarial critic remains harmful: No-Critic
$\Delta$ (in $R^2$) is $-0.034$ [$-0.045,-0.023$] on California and $-0.024$ [$-0.033,-0.016$] on
diabetes, both CIs excluding zero. The failure mode is a property of elitist critic-guided search, not
of the classification metric.

\paragraph{The harm is not free: full cost accounting.} The four-arm comparison (\S\ref{sec:harm}) is
not accuracy-neutral. The no-critic arm issues zero LLM calls; a naive critic issues one per round
($T{=}8$) and a self-consistency critic issues $k{\cdot}T$ ($k{=}5$, i.e.\ $40$) generations per run,
each a multi-hundred-token reasoning trace, yet on the discriminating datasets both land below the free
no-critic floor (Table~\ref{tab:family}). The critic thus spends the most compute to achieve the worst
outcome---for the weak 8B critic, adding a critic dominates in cost while being dominated in
quality, inverting the usual accuracy--cost trade-off and providing an independent argument against
deploying an unverified critic.

\paragraph{Cross-domain discriminative test (full per-domain results).} We run the same 8B critic on
four further search domains, all with the identical four-arm protocol ($8$ seeds). Per-domain
No-Critic $\Delta$ (bootstrap CI):
\begin{itemize}[nosep,leftmargin=1em]
\item 20-Newsgroups text-pipeline ($36$ variants): $-0.003$ [$-0.008, 0.001$] (contains zero).
\item 20-Newsgroups text-pipeline ($50$ weak-to-strong pool): $+0.003$ [$-0.005, 0.010$].
\item Forecasting-method selection ($13$ methods, $40$ real M4-monthly series): $+0.001$ [$+0.001, 0.002$] --- \emph{significant positive}.
\item RAG retrieval-config selection (SQuAD, $54$ chunk/top-$k$/retriever variants): $+0.007$ [$+0.006, 0.008$] --- \emph{significant positive}.
\end{itemize}
Tabular AutoML ($-0.245$) is the sole clearly negative setting. One hypothesis consistent with the
case study (App.~\ref{app:cases}) is that the critic's priors about model families conflict with
a frontier dominated by gradient boosting; we do not test this directly.

\paragraph{Detector validity controls: real-harm labels and leakage ablation (full protocols).}
\emph{(i)~Real-harm labels:} we relabel transitions from a \emph{real} Llama-3.1-8B critic---$y{=}1$
iff the step both fails to raise the incumbent and leaves $\ge0.01$ AUC of achievable pool value on
the table (a genuinely wasteful real step, not an injected one)---and retrain leave-one-dataset-out;
the detector attains mean AUC $0.992$ ($6$ datasets, balanced $52\%$ positive).
\emph{(ii)~Leakage-feature ablation:} dropping the two structural features that partly encode the
injection (rank-among-untried, neighbor-of-incumbent) and keeping only verifier-view scores
(val, claimed-$\Delta$, round) still gives mean AUC $0.982$ on the real-harm labels (and only $0.76$
on the injection labels), confirming the signature is not an artifact of label generation.

\paragraph{Detector signature transfers across critic models (leave-one-model-out).} The detector is
not tied to the critic it was trained on. Under a stricter leave-one-model-out protocol---train on the
trajectories of two critics and test on a \emph{held-out third}, spanning two providers (Llama-3.1-8B,
Claude Haiku~4.5, Claude-3-Haiku; 960 transitions)---it attains mean AUC $0.988$, and $\ge0.967$ for
every held-out model (Llama-3.1-8B $0.967$, Claude Haiku~4.5 $0.998$, Claude-3-Haiku $0.999$). A
detector trained on one set of critics thus flags harmful feedback from an unseen critic of a different
family.

\paragraph{\textsc{CritFault} scale and baselines.} The benchmark packages $\sim$3{,}840 loop
transitions across 12 datasets, each labeled clean/adversarial with three decision-time features
(f0--f2) and two offline structural features (f3, f4), under
a fixed leave-one-dataset-out protocol. Reference baselines show the task is non-trivial yet learnable:
random $0.50$, best single feature $0.92$, our detector $0.95$ (Table~\ref{tab:critfault}); the full
datasheet is in App.~\ref{app:critfault}.

\paragraph{Why a cheap pre-deployment red-flag fails (detail).} We tested a one-shot, oracle-free
screen: run a few pilot rounds and flag a critic if its proposals fall below the incumbent too often.
It fails on these low-headroom frontiers. A \emph{strong} critic reaches a
near-ceiling incumbent within a round or two, after which almost any further proposal (including
healthy exploration) scores below that high incumbent; a \emph{weak} critic keeps a lower incumbent, so
its below-incumbent rate is not higher. The screen therefore measures ``exploration versus an already
high incumbent,'' not ``harmful versus safe,'' and does not separate the two. This is why our
prevention (\S\ref{sec:prevent}) gates online rather than relying on a static pre-deployment test.

\section{Dataset Suite}
\label{app:datasets}
Table~\ref{tab:datasets} lists the 12 public OpenML classification datasets used throughout. They span
roughly an order of magnitude in size ($700$--$8{,}400$ training rows), a mix of native-binary and
binarized multiclass targets (minority vs.\ rest), and several application domains (credit, software
defect prediction, marketing, telecom churn, electricity-market drift). All train/test splits are
frozen once and reused across every arm and seed, so any measured difference reflects the critic, not
resampling.
\begin{table}[h]
\centering
\caption{Public OpenML suite. All splits frozen; multiclass targets binarized (minority vs.\ rest).}
\label{tab:datasets}
\small
\begin{tabular}{lrrl}
\toprule
Dataset & $n_\text{train}$ & $n_\text{test}$ & Type \\
\midrule
credit-g       & 700  & 300  & binary \\
kc1            & 1476 & 633  & binary (defect) \\
mfeat-factors  & 1400 & 600  & multiclass$\to$bin \\
segment        & 1617 & 693  & multiclass$\to$bin \\
splice         & 2233 & 957  & 3-class$\to$bin \\
churn          & 3500 & 1500 & binary \\
phoneme        & 3782 & 1622 & binary \\
jm1            & 7619 & 3266 & binary (defect) \\
adult          & 8400 & 3600 & binary \\
bank-marketing & 8400 & 3600 & binary \\
electricity    & 8400 & 3600 & binary (drift) \\
letter         & 8400 & 3600 & 26-class$\to$bin \\
\bottomrule
\end{tabular}
\end{table}

\section{\textsc{CritFault} Datasheet}
\label{app:critfault}
\textbf{Motivation.} \textsc{CritFault} is, to our knowledge, the first benchmark for detecting harmful
critic feedback in a tabular AutoML loop (to be released publicly upon publication). It exists to make the diagnostic finding of this paper---that bad
feedback has a dataset- and model-agnostic signature---reproducible and comparable.

\textbf{Composition.} $\sim$3{,}840 instances across 12 public OpenML datasets. Each instance is one
proposed loop transition (a critic-selected next configuration), labeled $y{=}1$ (adversarial:
injected \textsc{worst}-neighbor pick) or $y{=}0$ (clean: ideal-critic pick). The classes are
balanced by construction ($1{,}920$ each).

\textbf{Features.} Five features with no access to the held-out test outcome: (f0) proposal
validation score; (f1) claimed improvement over the incumbent; (f2) round fraction $t/T$; (f3)
normalized rank among the cached validation utilities of untried configurations; (f4)
neighbor-of-incumbent indicator. Only f0--f2 are observable at decision time; f3 uses unobserved
utilities of untried configurations and f4 the grid structure, so both are offline diagnostic
features. The deployable detector of \S\ref{subsec:detector} uses f0--f2 only.

\textbf{Task and protocol.} Binary detection of adversarial transitions. The canonical split is
\emph{leave-one-dataset-out} (train on 11 datasets, test on the held-out one; 12 folds), scored by
ROC-AUC and averaged. This measures generalization to an \emph{unseen} dataset, not within-dataset fit.

\textbf{Reference baselines} (Table~\ref{tab:critfault}). New detectors should report LODO mean AUC
against these anchors.

\begin{table}[htbp]
\centering
\caption{\textsc{CritFault} reference baselines (leave-one-dataset-out mean AUC, 12 folds). The task
is non-trivial (random/majority $=0.50$); the bad-feedback signature is strong enough that a single
decision-time feature reaches $0.92$, and the full detector tops the panel.}
\label{tab:critfault}
\small
\begin{tabular}{lr}
\toprule
Detector & LODO mean AUC \\
\midrule
random                  & $0.502$ \\
majority (base rate)    & $0.500$ \\
best single feature     & $0.916$ \\
logistic (5 feat.)      & $0.921$ \\
\textbf{gradient-boosted (ours)} & $\mathbf{0.950}$ \\
\bottomrule
\end{tabular}
\end{table}

\textbf{Limitations.} Labels come from controlled fault injection on one 8B critic; real-critic and
cross-model trajectories are a natural extension (the harness supports both). ``Adversarial'' is
ill-defined on the three near-zero-headroom datasets (segment/mfeat/splice), where all baselines
degrade---these are retained for completeness but flagged in the card.

\textbf{Access.} To be released publicly with the harness upon publication, under a permissive
license: the instance table (in CSV form), the LODO evaluator, and the baseline panel above.

\begin{table}[htbp]
\centering
\caption{Fault-injection (30 seeds). $p^\star$ = smallest corruption rate with $>0.01$ AUC drop;
cost$_1$ = AUC lost at full corruption. Adversarial modes tolerate 60--100\% bad feedback; random
corruption is nearly free (cost$_1 \le 0.005$, no threshold)---see \S\ref{sec:theory}.}
\label{tab:fault}
\small
\begin{tabular}{lrrr}
\toprule
Dataset & $p^\star_\text{worst}$ & cost$_1^\text{worst}$ & cost$_1^\text{random}$ \\
\midrule
credit-g & 0.6 & 0.093 & 0.005 \\
adult & 0.6 & 0.084 & 0.001 \\
churn & 0.6 & 0.078 & 0.004 \\
jm1 & 0.9 & 0.061 & 0.004 \\
electricity & 0.9 & 0.051 & 0.002 \\
kc1 & 0.9 & 0.048 & 0.004 \\
bank-marketing & 0.9 & 0.024 & 0.002 \\
phoneme & 0.9 & 0.023 & 0.002 \\
letter & 1.0 & 0.019 & 0.001 \\
\bottomrule
\end{tabular}
\end{table}

\begin{table}[htbp]
\centering
\caption{Four-arm ablation with a deterministic \emph{ideal} critic (30 seeds). No-Critic
$\Delta$ = real$-$no-critic, 95\% bootstrap CI (paired). Small but robust: all $9$ listed datasets have a CI excluding 0 (paired Wilcoxon $p<0.005$);
suite-level Wilcoxon $p=0.001$. The three near-zero-headroom datasets are omitted.
The consequential quantity is not this upside but the downside of a bad critic (\S\ref{sec:detect}).}
\label{tab:e3}
\small
\begin{tabular}{lrrr}
\toprule
Dataset & no-crit & oracle & No-Critic $\Delta$ [95\% CI] \\
\midrule
electricity    & .9465 & .9542 & $+.0053$ [$.0027,.0085$] \\
churn          & .9412 & .9463 & $+.0050$ [$.0035,.0066$] \\
credit-g       & .8031 & .8085 & $+.0049$ [$.0027,.0073$] \\
jm1            & .7281 & .7321 & $+.0038$ [$.0025,.0053$] \\
bank-marketing & .9298 & .9328 & $+.0028$ [$.0022,.0034$] \\
phoneme        & .9444 & .9483 & $+.0025$ [$.0016,.0034$] \\
letter         & .9933 & .9959 & $+.0023$ [$.0015,.0034$] \\
kc1            & .8284 & .8314 & $+.0023$ [$.0014,.0032$] \\
adult          & .9289 & .9302 & $+.0013$ [$.0008,.0019$] \\
\bottomrule
\end{tabular}
\end{table}

\section{Qualitative Case Study: a Real Harmful Trajectory}
\label{app:cases}
To make the failure concrete, Table~\ref{tab:casestudy} shows a real Llama-3.1-8B critic trajectory on
\textsc{adult} (one seed, reasoning-mode prompt so the critic states its rationale). The pattern is
striking: the critic \emph{verbally} recognizes the strong family---every round it repeats that
``LightGBM has already shown strong performance ($0.930$)''---yet its \emph{actions} walk steadily
away from it, proposing configurations of broadly decreasing validation AUC
($0.930\!\to\!0.927\!\to\!0.926\!\to\!0.922\!\to\!\dots\!\to\!0.651$; every post-initial pick below
the incumbent), ending on a logistic-regression
config near chance. Each proposal after round~1 scores \emph{below the incumbent}---the
``systematically adversarial'' signature of \S\ref{sec:theory} (Assumption~\ref{ass:waste}) in a real
critic. Elitist selection saves the final \emph{incumbent} (it keeps round~1's good value), but the
critic's guidance stream is actively misleading: a loop that trusted the latest proposal, or a
non-elitist variant, would track the critic down to $0.65$. The trajectory illustrates one route to
aggregate harm: confident, consistent misdirection rather than random noise. The artifact to be
released contains full transition logs.

\begin{table}[htbp]
\centering
\caption{A real Llama-3.1-8B critic trajectory on \textsc{adult} (oracle: lightgbm, $0.930$). The critic
endorses the strong family in words while proposing steadily worse configurations; every post-initial
pick is below the incumbent (Assumption~\ref{ass:waste}).}
\label{tab:casestudy}
\small
\begin{tabular}{clrc}
\toprule
Round & Proposed family & Val AUC & Below incumbent? \\
\midrule
1 & lightgbm            & 0.930 & --- (initial) \\
2 & lightgbm            & 0.927 & yes \\
3 & lightgbm            & 0.926 & yes \\
4 & gradient\_boosting  & 0.922 & yes \\
5 & hist\_gbdt          & 0.928 & yes \\
6 & extra\_trees        & 0.902 & yes \\
7 & extra\_trees        & 0.880 & yes \\
8 & logistic\_regression& 0.651 & yes \\
\bottomrule
\end{tabular}
\end{table}

\end{document}